\documentclass[10pt,conference,letterpaper]{IEEEtran}
\IEEEoverridecommandlockouts

\usepackage{amsmath,amssymb,amsthm}

\makeatletter
\def\th@plain{%
  \thm@notefont{}%
  \itshape 
}
\def\th@definition{%
  \thm@notefont{}%
  \normalfont 
}
\makeatother

\newtheoremstyle{boldstyle}
  {\topsep} 
  {\topsep} 
  {\itshape} 
  {} 
  {\bfseries} 
  {.} 
  {.5em} 
  {} 

\theoremstyle{boldstyle}
\newtheorem{theorem}{Theorem}
\newtheorem{lemma}[theorem]{Lemma}
\newtheorem{proposition}[theorem]{Proposition}

\newtheorem{remark}[theorem]{Remark}
\newtheorem{assumption}{Assumption}
\newtheorem{corollary}{Corollary}

\DeclareMathOperator{\Unif}{Unif}
\DeclareMathOperator{\diam}{diam}
\DeclareMathOperator{\SPD}{SPD}
\DeclareMathOperator{\Risk}{Risk}
 
\DeclareMathOperator{\Dec}{Dec} 
\DeclareMathOperator{\Enc}{Enc}

\newcommand{\Gcal}{\mathcal{G}}
\newcommand{\Ccal}{\mathcal{C}}
\newcommand{\Tcal}{\mathcal{T}}
\newcommand{\Fcal}{\mathcal{F}}
\newcommand{\Nb}{\mathcal{N}}
\newcommand{\Nbb}{\mathbb{N}}
\newcommand{\Rbb}{\mathbb{R}}
\newcommand{\bd}{\mathbf{d}}
\newcommand{\E}{\mathbf{E}}
\newcommand{\R}{\mathbf{R}}

\usepackage{siunitx}
\usepackage{graphicx}
\usepackage{caption}
\usepackage{subcaption}
\usepackage{float}
\usepackage{booktabs}
\usepackage{multirow}
\usepackage{tabularx}
\usepackage{colortbl}
\usepackage{xcolor}
\usepackage{textcomp}
\usepackage{afterpage}
\usepackage{gensymb}
\usepackage{enumitem}
\usepackage{array}
\usepackage{algorithm}
\usepackage[noend]{algpseudocode}
\usepackage{cite}
\newcolumntype{L}[1]{>{\raggedright\arraybackslash}m{#1}}
\newcolumntype{C}[1]{>{\centering\arraybackslash}m{#1}}
\newcolumntype{R}[1]{>{\raggedleft\arraybackslash}m{#1}}

\algrenewcommand\algorithmicrequire{\textbf{Require:}}
\algrenewcommand\algorithmicensure{\textbf{Ensure:}}

\usepackage[hidelinks]{hyperref}  

\hypersetup{
    colorlinks=true,
    linkcolor=black,
    citecolor=blue,
    urlcolor=blue,
    pdfborder={0 0 0}  
}

\def\BibTeX{{\rm B\kern-.05em{\sc i\kern-.025em b}\kern-.08em
  T\kern-.1667em\lower.7ex\hbox{E}\kern-.125emX}}

\title{
Resolving Node Identifiability in Graph Neural Processes via Laplacian Spectral Encodings
\thanks{
This work was supported by the National Natural Science Foundation of China [61773020] and the Graduate Innovation Project of National University of Defense Technology [XJQY2024065]. The authors would like to express their sincere gratitude to all the referees for their careful reading and insightful suggestions.
}
}

\author{%
{ Zimo Yan\textsuperscript{\rm 1}, Zheng Xie\textsuperscript{\rm 1}\thanks{*Corresponding author: Zheng Xie (xiezheng81@nudt.edu.cn).}, Chang Liu\textsuperscript{\rm 1}, Yuan Wang\textsuperscript{\rm 2} }%
\vspace{1.6mm}\\
\fontsize{10}{10}\selectfont\itshape
\textsuperscript{\rm 1}National University of Defense Technology, Changsha, China.\\
\textsuperscript{\rm 2}School of Urban Design, Wuhan University, Wuhan, China.
\fontsize{9}{9}\selectfont\ttfamily
\\\{yanzimo20, xiezheng81, nudt\_liuchang\_1997 \}@nudt.edu.cn, 2024282090042@whu.edu.cn}
\fontsize{9}{9}\selectfont\ttfamily\upshape
\fontsize{10}{10}\selectfont\rmfamily\itshape

\fontsize{9}{9}\selectfont\ttfamily\upshape

\begin{document}
\maketitle

\begin{abstract}
Message passing graph neural networks are widely used for learning on graphs, yet their expressive power is limited by the one-dimensional Weisfeiler–Lehman test and can fail to distinguish structurally different nodes.
We provide rigorous theory for a Laplacian positional encoding that is invariant to eigenvector sign flips and to basis rotations within eigenspaces. We prove that this encoding yields node identifiability from a constant number of observations and establishes a sample-complexity separation from architectures constrained by the Weisfeiler–Lehman test. The analysis combines a monotone link between shortest-path and diffusion distance, spectral trilateration with a constant set of anchors, and quantitative spectral injectivity with logarithmic embedding size. As an instantiation, pairing this encoding with a neural-process style decoder yields significant gains on a drug–drug interaction task on chemical graphs,When tested empirically on a challenging Drug-Drug Interaction (DDI) prediction task, improving both the area under the ROC curve and the F1 score and demonstrating the practical benefits of resolving theoretical expressiveness limitations with principled positional information.

\vspace{1em} 
\end{abstract}

\section{Introduction}

Learning effective representations from graph-structured data is a fundamental challenge with profound implications across numerous domains, from molecular chemistry and drug discovery \cite{gilmer2017neural} to social network analysis and recommendation systems \cite{ying2018graph}. Graph Neural Networks (GNNs) have emerged as the dominant paradigm for this task, employing an iterative message-passing scheme to capture local neighborhood structures \cite{kipf2017semi}. More recently, the integration of GNNs into probabilistic frameworks like Graph Neural Processes (GNPs) \cite{garnelo2018conditional, yan2025multiscalegraphneuralprocess} has unlocked new capabilities. By learning a distribution over functions on graphs, GNPs can not only make predictions but also quantify their own uncertainty, a feature that is invaluable for safety-critical applications and few-shot learning scenarios where data is scarce.

The predominant approach for constructing GNPs involves using a standard GNN, such as a Graph Convolutional Network (GCN) or GraphSAGE, as the primary encoder \cite{kim2019attentive}. These models, which we categorize as Weisfeiler-Lehman GNPs (WL-GNPs), first generate a deterministic embedding for each node and then aggregate them to produce a global representation of the graph. This representation is subsequently used by the neural process to condition its predictions. While this combination is powerful, the resulting model inherits all the intrinsic limitations of its underlying GNN encoder.

A well-documented deficiency of standard GNNs is that their expressive power is fundamentally bounded by the 1-Weisfeiler-Lehman (1-WL) test for graph isomorphism \cite{xu2019how, morris2019weisfeiler}. This theoretical ceiling means that WL-GNPs are incapable of distinguishing between certain non-isomorphic graph structures, a failure that is particularly pronounced in regular or highly symmetric graphs. Within the GNP framework, this limitation translates into a critical problem of \textit{node indistinguishability}. If the encoder cannot differentiate two nodes based on structure, it will produce identical representations, leading to ambiguous posteriors and a provably high Bayes risk for tasks that depend on identifying a specific node's role or position.

\textbf{Motivation}:
This fundamental limitation of standard models raises a critical research question: \textit{Can we design a Graph Neural Process that provably overcomes the expressive limitations of the 1-WL test, thereby resolving the issue of node indistinguishability and reducing the inherent Bayes risk?}

To address this challenge, we turn to the concept of positional encodings (PEs), which have been successfully used to inject positional awareness into permutation-invariant architectures like the Transformer \cite{dwivedi2020benchmarking}. Inspired by recent work leveraging the graph Laplacian spectrum to create robust node positions \cite{kreuzer2021rethinking}, we propose a new class of models, the Laplacian Graph Neural Process (Lap-GNP). Instead of relying on flawed local structures, Lap-GNP equips each node with a sign-invariant positional encoding derived from the Laplacian eigenvectors, providing a global "coordinate system". In this work, we provide a rigorous theoretical proof that this principled approach allows Lap-GNP to achieve constant-shot identifiability on random regular graphs, a setting where we prove WL-GNPs are guaranteed to fail. We then empirically validate our theory on a challenging Drug-Drug Interaction (DDI) prediction task, demonstrating that the theoretical boost in expressiveness translates directly into significant gains in performance and sample efficiency.

\textbf{Outline \& Primary Contributions}:  
This paper is organized as follows. Section~\ref{SE2} provides a review of related work on graph neural networks and positional encodings.
Section~\ref{sec:preliminaries} introduces the necessary preliminaries, notations, and problem formulation.  
Section~\ref{SE5} presents our core theoretical analysis, establishing the lower bound for WL-GNPs and the upper bound for Lap-GNPs.  
Section~\ref{SE7} describes the experimental setup and presents empirical results and provides an in-depth discussion of model behavior, connecting our theoretical findings to practical performance. 
Section~\ref{SE9} concludes the paper and outlines potential directions for future work.  

The primary contributions of this work are as follows:
\begin{enumerate}
     \item We formally characterize the failure of standard GNN-based encoders (WL-GNPs) within a probabilistic setting, proving they suffer from a high Bayes risk due to node indistinguishability on symmetric graphs.
     \item We propose the Laplacian Graph Neural Process (Lap-GNP), a novel model class that incorporates a sign-invariant spectral positional encoding to resolve structural ambiguity. We prove a formal sample-complexity separation, showing Lap-GNP achieves constant-shot identifiability where WL-GNPs fail.
     \item We provide compelling empirical validation on a real-world Drug-Drug Interaction (DDI) prediction task, demonstrating that our theoretically-grounded Lap-GNP significantly outperforms standard baselines, thus bridging the gap between expressiveness theory and practical application.
\end{enumerate}

\section{Literature Review}
\label{SE2}

\subsection{Graph Representation Learning on Graph-Structured Data}
\label{sec:lr_graph}
Learning on graphs underpins node-, edge-, and graph-level tasks, including structural role discovery, link prediction, and molecular property prediction \cite{gilmer2017neural,hamilton2017inductive,wu2020comprehensive}. Message-Passing Graph Neural Networks (MPNNs) have become the prevailing paradigm: nodes iteratively aggregate neighborhood information to form task-specific representations \cite{gilmer2017neural,xu2019how}. Representative instances span GCN \cite{kipf2017semi}, GraphSAGE \cite{hamilton2017inductive}, GAT \cite{velickovic2018graph}, and GIN \cite{xu2019how}, with numerous architectural refinements \cite{dwivedi2020benchmarking,corso2020pna}. Despite empirical success, three challenges recur. 
First, expressiveness is bounded by the 1-Weisfeiler–Lehman (1-WL) test \cite{xu2019how,morris2019weisfeiler}, which renders many non-isomorphic but structurally similar subgraphs indistinguishable. 
Second, deeper stacks suffer \emph{over-smoothing}, where embeddings homogenize across large regions \cite{li2018deeper,oono2020graph}. 
Third, long-range dependencies are compressed through narrow channels, leading to \emph{over-squashing} \cite{alon2021bottleneck,topping2022understanding}. 
Remedies include higher-order or subgraph variants \cite{morris2019weisfeiler,bevilacqua2022equivariant}, diffusion-augmented propagation \cite{klicpera2019predict,klicpera2019diffusion}, and graph transformers that relax strict locality via attention \cite{ying2021graphormer,rampasek2022gps,kreuzer2021rethinking}. These often trade accuracy for efficiency on sparse graphs unless complemented with principled positional priors.

\subsection{Neural Processes and Graph Neural Processes}
\label{sec:lr_np_gnp}
Neural Processes (NPs) model distributions over functions conditioned on context observations, combining amortized inference with GP-like uncertainty \cite{garnelo2018conditional,kim2019attentive}. Variants expand representation capacity via attention and hierarchical latents \cite{kim2019attentive,lee2020bootstrapping}. On graphs, Graph Neural Processes (GNPs) instantiate NP encoders with GNN backbones to capture relational structure, enabling few-shot generalization and calibrated predictive uncertainty in relational domains. In practice, most GNPs inherit the expressiveness ceiling of their underlying MPNN encoders and thus exhibit posterior ambiguity on symmetric graphs due to 1-WL limits \cite{xu2019how,morris2019weisfeiler}. Complementary probabilistic graph models, such as VAEs on graphs \cite{yan2025metamolgenneuralgraphmotif,kipf2016variational} and Bayesian GNNs \cite{hasanzadeh2020bayesian}, underscore the value of uncertainty for reliable decision-making, yet few works connect positional design to probabilistic \emph{identifiability}. Our companion theory formalizes this gap by establishing a Bayes-risk lower bound for WL-bounded GNPs and a sample-complexity separation once sign- and basis-invariant Laplacian positional information is introduced into the encoder.

\subsection{Positional Encodings for Graphs: Structural, Diffusive, and Spectral}
\label{sec:lr_pe}
Graph positional encodings (PEs) endow permutation-invariant learners with position awareness. 
\emph{Structural} PEs encode discrete distances or generalized diffusion scores (e.g., shortest paths, PPR), improving substructure discrimination beyond purely local message passing \cite{dwivedi2020benchmarking}. 
\emph{Diffusive} PEs arise from random walks or heat kernels and emphasize multiscale propagation patterns \cite{klicpera2019diffusion,giuliari2024positional}. 
\emph{Spectral} PEs embed nodes via functions of Laplacian eigenpairs, from truncated eigenvectors to full-spectrum parameterizations \cite{dwivedi2020benchmarking,kreuzer2021rethinking,ying2021graphormer,rampasek2022gps}. Spectral designs can break graph symmetries and strengthen expressiveness when coupled with attention; however, raw eigenvectors exhibit sign ambiguity and basis rotations within eigenspaces. Recent advances impose \emph{sign- and basis-invariance} at the feature level \cite{lim2022signnet}, and integrate PEs either as augmented node features or as aggregation controllers \cite{dwivedi2020benchmarking,kreuzer2021rethinking}. In our setting, we adopt a Laplacian PE that is invariant to these ambiguities and prove that, under mild spectral injectivity and local treelike conditions, it restores \emph{identifiability} within a GNP encoder.

\subsection{Limitations and Gap Analysis}
\label{sec:lr_gap}
Two gaps emerge from the literature. 
\emph{Theory.} GNPs built on WL-bounded encoders inherit the 1-WL ceiling \cite{xu2019how,morris2019weisfeiler}, which induces posterior ambiguity among structurally indistinguishable nodes in symmetric graphs. PE studies rarely connect spectral invariance choices to probabilistic identifiability or Bayes-risk guarantees \cite{lim2022signnet}. Our analysis closes this gap by showing WL-GNPs have a nontrivial Bayes-risk lower bound under sublogarithmic context size, whereas Laplacian-enhanced GNPs achieve constant-shot identifiability under sign- and basis-invariant spectral encodings and logarithmic depth budgets. 
\emph{Practice.} Many graph systems lack context-aware modeling and calibrated uncertainty in downstream applications. We address this gap in Drug–Drug Interaction (DDI) prediction by adopting a strong GNP-based baseline and showing consistent improvements once principled positional information is incorporated, aligning theory with empirical reliability.

\subsection{DDI Prediction: From Similarity and Dual-GNN Pipelines to Probabilistic, Context-Aware Models}
\label{sec:lr_ddi}
DDI prediction has evolved from literature mining and similarity-based models \cite{percha2013profiling,gottlieb2012ddi,ryu2018deepddi} to graph-centric approaches. A landmark graph method, Decagon, embeds drugs in heterogeneous biomedical networks for multi-relational prediction \cite{zitnik2018decagon}, while dual-GNN pipelines encode paired drugs independently before late fusion \cite{ma2023dualgnn,liu2021comparative}. Recent advances integrate knowledge graphs and multimodal signals \cite{lin2020kgddi,li2024multimodal}, and introduce co-attention or gated mechanisms to capture substructure-level interactions \cite{Nyamabo2021ssidddi, yan2025multiscalegraphneuralprocess}. Yet many models produce static, context-agnostic embeddings and lack calibrated uncertainty. 
We adopt a multi-scale GNP with cross-drug co-attention as our baseline configuration, aligned with recent probabilistic formulations, datasets, and protocols used in DDI benchmarks \cite{law2014drugbank,Shen2025baseline}. This setting grounds our empirical comparison and highlights the added value of theoretically principled positional information within GNPs for DDI tasks.

\subsection{Relation to Distance Encoding (DE) and the role of \texorpdfstring{$p$}{p}}
\label{sec:de-background}
\begin{figure}[htbp]
    \centering
    \includegraphics[width=\linewidth]{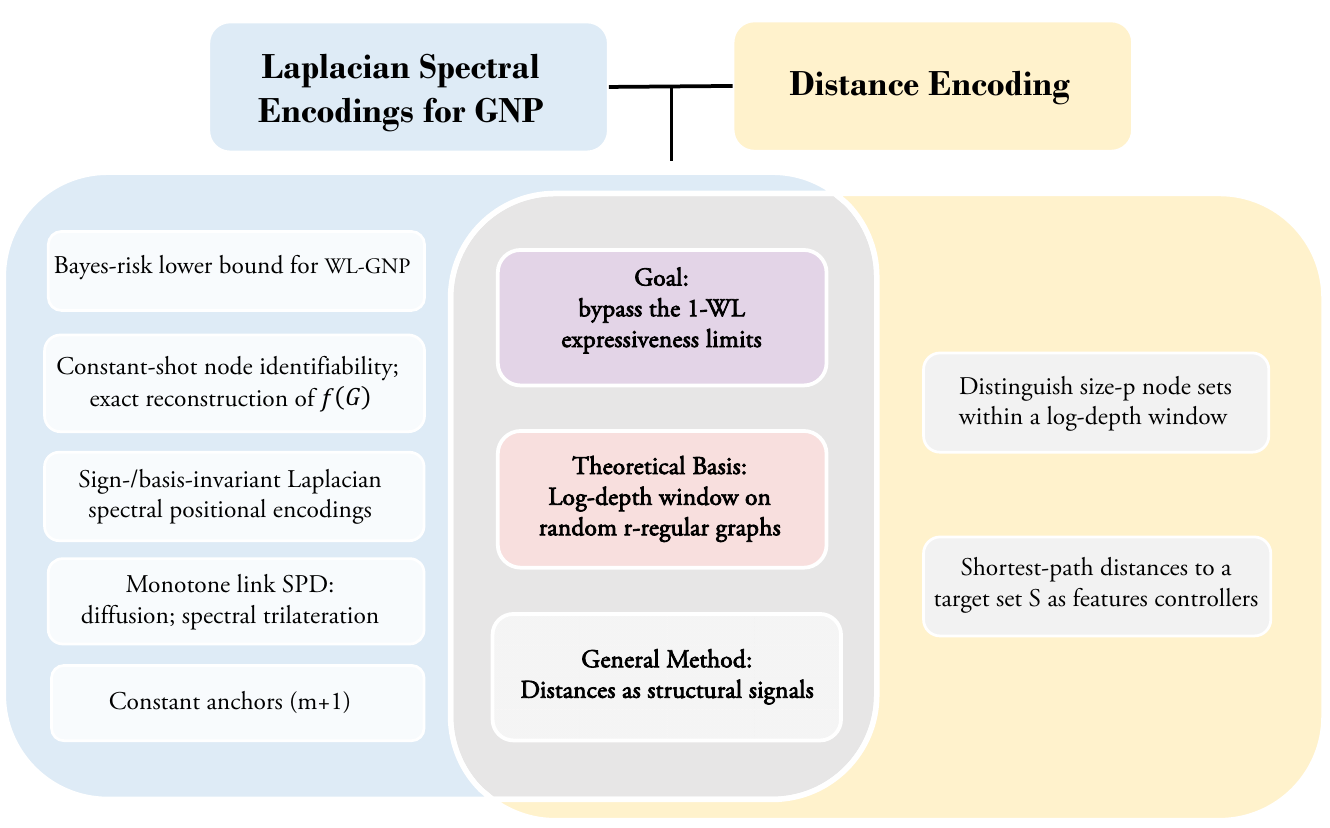}
    \caption{Venn-style overview of our theory versus DE. The overlap shows shared assumptions/goals (log-depth window on random $r$-regular graphs; distances as structural signals; going beyond 1-WL), while the left/right lobes list theory unique to our anchor-based Laplacian spectral approach and to DE, respectively.}
    \label{fig:theorem-venn}
\end{figure}

 Figure~\ref{fig:theorem-venn} provides a one-glance summary; we now formalize the relation to DE and the role of $p$. Distance Encoding (DE) \cite{li2020distance} constructs a permutation-invariant distance feature $\zeta(\cdot\,|\,S)$ for a target node set $S$ and denotes by \emph{DE-$p$} the case $|S|=p$ (see \cite[Def.~3.1]{li2020distance}). Their main theorem shows that, on (almost all) random $r$-regular graphs, a DE-GNN-$p$ can distinguish two size-$p$ node sets within the logarithmic depth window
\begin{align}
L \;\le\; \Bigl(\tfrac{1}{2}+\varepsilon\Bigr)\,\frac{\log n}{\log(r-1)} ,
\end{align}
while \emph{DE-1} exhibits limitations on distance-regular graphs \cite[Thm.~3.7, Cor.~3.8]{li2020distance}.

Our task is \emph{single-source node identifiability} ($p_{\text{task}}=1$). Instead of directly encoding the target set as in DE, we introduce a \emph{constant-size} set of \emph{anchors} $S=\{a_i\}_{i=1}^{m+1}$ (so $p_{\text{anchor}}=m{+}1$) and only query shortest-path distances from these anchors to the hidden source. Conceptually, the anchors act as a reference frame: observed shortest-path signals are passed through a monotone distance–to–diffusion linkage and then used to perform a small-dimensional geometric reconstruction (trilateration) of the source’s position in a robust spectral coordinate system. This \emph{anchor-based} mechanism operates under the same logarithmic depth window as DE’s main theorem but targets $p_{\text{task}}{=}1$ with a \emph{fixed} anchor cardinality. The formal realization of this idea—via tree monotonicity with random-regular stability, spectral trilateration, and quantitative spectral injectivity—is established later in our main theorem (Sec.~\ref{subsec:proof-main}).

\section{Preliminaries}
\label{sec:preliminaries}

\subsection{Graphs and Notations}
\label{subsec:notation}

Let $G=(V,E)$ be a finite, simple, connected, undirected graph with $|V|=n$.
We write $A\in\{0,1\}^{n\times n}$ for the adjacency matrix, $D$ for the diagonal degree matrix,
and define the (symmetric) normalized Laplacian
\begin{align}
L_{\mathrm{sym}} \;:=\; I - D^{-1/2} A D^{-1/2}.
\end{align}
We also use the random-walk Laplacian $L_{\mathrm{rw}}:=I-P$ with $P:=D^{-1}A$.
On $r$-regular graphs, $D=rI$ so
\begin{align}
L_{\mathrm{sym}} \;=\; L_{\mathrm{rw}} \;=\; I - \tfrac{1}{r}A,
\end{align}
hence throughout our tree/regular-graph arguments we identify
\begin{align}
L \;:=\; I - \tfrac{1}{r}A \qquad\text{(symmetric).}
\end{align}

For $u,v\in V$, let $\SPD_G(u,v)$ (or simply $\SPD(u,v)$ when unambiguous) denote the shortest-path distance, and $\Nb(u)$ the neighbor set of $u$.
For $u\in V$ and $R\in\mathbb{N}$, let the radius-$R$ ball be
\begin{align}
B_G(u,R) \;:=\; \{\,v\in V:\ \SPD_G(u,v)\le R\,\}.
\end{align}
We write $\diam(G)$ for the diameter.
We use “w.h.p.” to mean probability $1-o(1)$ as $n\to\infty$, and $\Gcal_{n,r}$ for the uniform distribution on labeled $r$-regular graphs on $[n]$.
We write $\log n:=\log_2 n$ and $(\log n)^{O(1)}$ for a polylogarithmic factor.
If $R$ is a random object (e.g., an encoder output), $\sigma(R)$ denotes the $\sigma$-field it generates.

\paragraph{Heat kernel and diffusion distance.}
For $t>0$, define the heat semigroup and kernel
\begin{align}
K_t \;:=\; e^{-tL}, \qquad k_t(u,v) \;:=\; (K_t)_{uv}.
\end{align}
Let $\{(\lambda_j,\phi_j)\}_{j\ge1}$ be the nonzero eigenpairs of $L$ with $\{\phi_j\}$ orthonormal in $\ell^2(V)$.
The (full) diffusion distance is
\begin{align}\label{eq:dt-prelim}
d_t(u,v)^2 \;=\; \sum_{j\ge1} e^{-2t\lambda_j}\big(\phi_j(u)-\phi_j(v)\big)^2.
\end{align}
Using the spectral expansion of $K_t$,
\begin{align}
K_t \;=\; \sum_{j\ge1} e^{-t\lambda_j}\,\phi_j\phi_j^\top,
\end{align}
one obtains the kernel identity
\begin{align}\label{eq:dt-kernel-id}
d_t(u,v)^2 \;=\; k_{2t}(u,u)+k_{2t}(v,v)-2k_{2t}(u,v).
\end{align}
On $r$-regular graphs (and on vertex-transitive graphs like the infinite $r$-regular tree), $k_{2t}(u,u)$ is constant in $u$; this yields the simplified form later used in our tree-based arguments.
When $L=I-\tfrac{1}{r}A$, we also use the Poissonization identity
\begin{align}
K_t \;=\; e^{-t}\,e^{\,tP}, \qquad P:=\tfrac{1}{r}A,
\end{align}
which is convenient for comparing $G$ with its tree cover within logarithmic radii.

\paragraph{Affine independence and vectorization.}
Points $p_1,\dots,p_{m+1}\in\R^m$ are \emph{affinely independent} if $\{p_i-p_{m+1}\}_{i=1}^m$ are linearly independent in $\R^m$.
For a symmetric matrix $M$, $\mathrm{vec}_\triangle(M)$ stacks the strict upper-triangular entries of $M$ in a fixed order.
We use $\mathrm{concat}(\cdot)$ for coordinate-wise concatenation.

\subsection{GNP Task and Model Classes}
\label{subsec:gnp_models}

\textbf{GNP Task.}
Fix a graph $G$ and an unobserved source $v_0\sim\Unif(V)$.
Define
\begin{align}
f_{G,v_0}(v) \;=\; \SPD(v,v_0).
\end{align}
The learner receives a context set
\begin{align}
\Ccal \;=\; \{(v_i,y_i)\}_{i=1}^k, v_i\stackrel{\text{i.i.d.}}{\sim}\Unif(V),\quad y_i=f_{G,v_0}(v_i).
\end{align}
A Graph Neural Process (GNP) has an encoder $R=\Enc(G,\Ccal)$ and a decoder $\widehat f(t)=\Dec(t,R)$.

\begin{figure*}[htbp]
    \centering
    \includegraphics[width=0.8\textwidth]{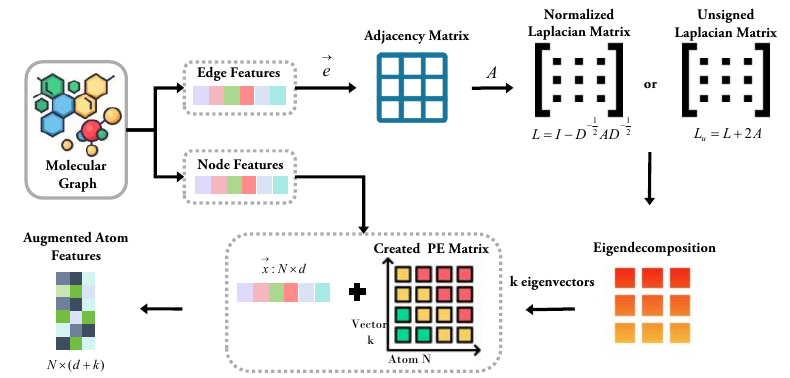}
    \caption{\textbf{The Laplacian Positional Encoding (PE) pipeline.}
    For a molecular graph with $N$ atoms, its adjacency matrix is used to compute a Laplacian matrix ($L$ or $L_u$).
    The first $k$ eigenvectors of this Laplacian are extracted to form an $N\times k$ PE matrix.
    Finally, each atom's original $d_{\text{node}}$-dimensional feature vector is concatenated with its corresponding $k$-dimensional PE vector, resulting in an augmented $N\times(d_{\text{node}}+k)$ feature matrix for encoder.}
    \label{fig:pe_pipeline}
\end{figure*}

\textbf{Model Classes.}
(i) \emph{WLGNP}: encoders whose message-passing power is bounded by the $1$-WL test (e.g., MPNNs), hence with trivial initial features they inherit $1$-WL limitations~\cite{morris2019weisfeiler,xu2019how}.\\
(ii) \emph{Lap-GNP}: the same backbone augmented with a sign-/basis-invariant spectral positional encoding $\Psi(v)$ built from the first $M$ nonzero eigenpairs $\{(\lambda_j,\phi_j)\}_{j=1}^M$ of $L$; our practical pipeline follows Fig.~\ref{fig:pe_pipeline}.

\textbf{Spectral PE (sign-/basis-invariant).}
Let
\begin{align}\label{eq:phi-stack}
\phi^{(M)}(v) \;=\; (\phi_1(v),\dots,\phi_M(v))^\top \in\R^M.
\end{align}
Define
\begin{align}\label{eq:s-vec}
s(v) \;=\; \phi^{(M)}(v)\odot\phi^{(M)}(v) \;=\; (\phi_1(v)^2,\dots,\phi_M(v)^2)\in\R^M,
\end{align}
and
\begin{align}
u(v) &=\mathrm{vec}_\triangle\!\Big(\,\big|\phi^{(M)}(v)\,{\phi^{(M)}(v)}^\top\big|\,\Big) \\
&= \big(|\phi_j(v)\phi_{j'}(v)|\big)_{1\le j<j'\le M}\in\R^{\frac{M(M-1)}{2}}.
\end{align}
Set
\begin{align}\label{eq:psi}
\Psi(v) \;=\; \mathrm{concat}\!\Big(\{\lambda_i\}_{i=1}^M,\ s(v),\ u(v)\Big)\in\R^{\,M+M+\frac{M(M-1)}{2}}.
\end{align}

\subsection{Bayes Risk Framework for Identifiability}
\label{subsec:bayes_framework}

To formalize “failure to identify,” let the prior on the hidden source be $\pi=\Unif(V)$ and the random observation be $Z:=(G,\Ccal)$.
A decision rule $\delta$ outputs $\widehat v=\delta(Z)$.
The $0$–$1$ Bayes risk is
\begin{align}
\Risk(\delta) \;:=\; \Pr\!\big[\delta(Z)\neq v_0\big].
\end{align}
For an encoder class $\Fcal$ producing a summary $R=R(G,\Ccal)$, define indistinguishability by
\begin{align}
u\sim_{\Fcal,Z} v \quad\Longleftrightarrow\quad \Pr(v_0=u\mid Z,R)=\Pr(v_0=v\mid Z,R).
\end{align}
Let $[u]=\{w\in V:\ w\sim_{\Fcal,Z} u\}$.
Any Bayes-optimal rule must choose $\widehat v\in[v_0]$, hence
\begin{align}\label{eq:riskbucket}
\Pr(\widehat v\ne v_0\mid Z,R)
\;\ge\; 1-\E\!\big[\,|[v_0]|^{-1}\,\bigm|\,Z,R\big].
\end{align}
Thus, showing that $|[v_0]|\ge 2$ w.h.p.\ yields a nontrivial lower bound on the Bayes risk.

\subsection{Why Random $r$-Regular Graphs}
\label{subsec:why_regular}

Random $r$-regular graphs $G\sim\Gcal_{n,r}$, where $\Gcal_{n,r}$ denotes the uniform distribution on labeled $r$-regular graphs on $[n]$ and the degree $r\ge 3$ is fixed, are a canonical hard family for $1$-WL/MPNNs with trivial features: they are highly symmetric and locally tree-like up to a logarithmic radius, and they constitute the standard setting in DE’s main theorem (their Thm.~3.3). We adopt the same depth window to state sharp depth/sample bounds in our analysis.

We deliberately develop the theory on random $r$-regular graphs because they realize the canonical \emph{hard case} for $1$-WL/MPNNs: symmetry is maximal and local neighborhoods are tree-like up to $\Theta(\log n)$. Proving constant-shot identifiability for Lap-GNP in this regime therefore serves as a worst-case guarantee. In non-regular, heterogeneous graphs, degree variability and structural asymmetries \emph{break} these symmetries, typically enlarging spectral separation and making identifiability only easier under the same anchor-based pipeline.

\section{Theoretical Analysis}
\label{SE5}

In this section, we formally prove the sample-complexity separation between WLGNPs and Lap-GNPs on $G\sim\Gcal_{n,r}$.
\subsection{Main Result}
In what follows, we first specify the structural assumptions and technical lemmas under which Theorem~\ref{thm:main} holds, and then show that these conditions are satisfied with high probability on random $r$-regular graphs. We finally combine these ingredients to give a complete proof of Theorem~\ref{thm:main}.
\begin{theorem}\label{thm:main}
Let $G\sim\Gcal_{n,r}$ with fixed $r\ge 3$ and $v_0\sim\Unif(V)$.
\begin{enumerate}
\item[(i)] (\emph{WLGNP lower bound}) If $k=o(\log n)$, any WLGNP has Bayes risk $\Risk(\delta)\ge\tfrac12(1-o(1))$ for identifying $v_0$.
\item[(ii)] (\emph{Lap-GNP constant-shot identifiability}) 
Assume \emph{Sign-/basis-invariance} (Assumption~\ref{assump:sign_invariant}), \emph{Spectral injectivity with separation} (Assumption~\ref{assump:injective}), \emph{Monotone linkage with stability} (Assumption~\ref{assump:monotone}, Lemma~\ref{lem:monotone}), and \emph{Anchor general position} (Assumption~\ref{assump:anchors}). 
Then there exist constants $M=\Theta(\log n)$, a time $t>0$, an integer $m\in\Nbb$, and $k_0=m+1$, such that any Lap-GNP with depth $L$ satisfying \eqref{eq:depth} and with $k\ge k_0$ i.i.d.\ context points identifies $v_0$ uniquely with probability $1-o(1)$, and consequently reconstructs $f_{G,v_0}$ exactly on all of $V$.
\end{enumerate}
\end{theorem}

\subsection{Setup and Assumptions}\label{subsec:assumptions}
Let $G\sim\Gcal_{n,r}$ with fixed $r\ge 3$, and let $D:=\diam(G)$. It is known that $D=\Theta(\log n)$ w.h.p.
We consider a context size $k=o(\log n)$. For any node $u\in V$, we define its \emph{context-distance vector} as
\begin{equation}
\bd(u)\ :=\ \bigl(\SPD(u,v_1),\dots,\SPD(u,v_k)\bigr)\ \in\ \{0,1,\dots,D\}^k.
\end{equation}
This vector induces a partition of the nodes into "buckets," where all nodes in a bucket share the same context-distance vector.

Throughout this paper let $r\ge 3$ be a fixed constant and $G\sim\mathcal G_{n,r}$ be a uniformly random labeled $r$-regular graph on $[n]$. For $u\in V(G)$ and $T\in\mathbb N$, write $B_G(u,T):=\{v:\ \SPD_G(u,v)\le T\}$ for the radius-$T$ ball.

There exists a constant $\epsilon_0>0$ such that for
\begin{align}\label{eq:T}
T\ \le\ \Bigl(\tfrac12-\epsilon_0\Bigr)\frac{\log n}{\log(r-1)},
\end{align}
the following hold:
\begin{itemize}
\item  If $u\sim\Unif(V)$ is sampled independently of $G$, then $\Pr\big(B_G(u,T)\text{ is a tree}\big)=1-o(1)$.
\item  With probability $1-o(1)$ over $G$, at most $o(n)$ vertices $u$ have $B_G(u,T)$ containing a cycle. Equivalently, $B_G(u,T)$ is a tree for all but $o(n)$ vertices $u$.
\end{itemize}
Fix any constant $c>0$. For depth
\begin{align}\label{eq:depth}
L\ \le\ \Biggl\lceil\Bigl(\tfrac12+c\Bigr)\frac{\log n}{\log(r-1)}\Biggr\rceil,
\end{align}
and for any \emph{finite or polylogarithmic} number of seed nodes (e.g., anchors and the hidden source), the encoder can, w.h.p.\ over $G$, stably propagate signals across $\Theta(\log n)$-radius neighborhoods around those seeds without strong cycle interference (same logarithmic window as DE Thm.~3.3).%
\footnote{At scale $T=\Theta(\log n)$, cycles of length $\Theta(\log n)$ do occur in $G$, so the treelike claim cannot hold for \emph{every} vertex simultaneously. The statements above are standard: the failure probability for one root is $O((r\!-\!1)^{2T}/n)$, hence $o(1)$ when $T\le(\tfrac12-\epsilon_0)\log n/\log(r-1)$; a union bound over $(\log n)^{O(1)}$ seeds still yields $o(1)$.}

\textbf{Spectral ingredients.}
Define the diffusion (heat-kernel) distance at time $t>0$:
\begin{align}\label{eq:dt}
d_t(u,v)^2  := \sum_{j\ge 1} e^{-2t\lambda_j}\bigl(\phi_j(u)-\phi_j(v)\bigr)^2 .
\end{align}
Let $\Phi_t(v):=(e^{-t\lambda_j}\phi_j(v))_{j=1}^m$ be the $m$-dimensional truncated spectral embedding.

We now state the structural assumptions under which we prove identifiability, and then argue that they hold w.h.p.\ in $G\sim\Gcal_{n,r}$. In adition, Assumption\ref{assump:injective} will be verified in Theorem\ref{thm:inject} and Corollary\ref{cor:verifyA2}

\begin{assumption}\label{assump:sign_invariant}
The node-wise positional encoding $\Psi$ defined in~\eqref{eq:psi} is invariant to eigenvector sign flips and orthogonal changes of basis within any eigenspace of $L$.
\end{assumption}

\begin{assumption}\label{assump:injective}
There exist $M=\Theta(\log n)$ and $\alpha>0$ such that, w.h.p.,
$\Psi:V\to\Rbb^d$ is injective and
\begin{align}\label{eq:min-sep}
\min_{u\ne v}\|\Psi(u)-\Psi(v)\|\ \ge\ n^{-\alpha}.
\end{align}
\end{assumption}

\begin{assumption}\label{assump:monotone}
There exist $t\in[t_-,t_+]$, $R=\Theta(\log n)$, $c_1,c_2>0$ and a strictly increasing $\psi:[0,R]\to\Rbb_+$ such that, w.h.p., for all $u,v$ with $\SPD(u,v)\le R$,
\begin{align}\label{eq:mono}
c_1\,\psi\!\big(\SPD(u,v)\big)\ \le\ d_t(u,v)\ \le\ c_2\,\psi\!\big(\SPD(u,v)\big).
\end{align}
\end{assumption}

\begin{assumption}\label{assump:anchors}
For some constant $m$ and $k_0=m+1$, $k_0$ i.i.d.\ uniform anchors $\{v_i\}$ satisfy that $\{\Phi_t^{(m)}(v_i)\}_{i=1}^{k_0}$ are affinely independent in $\Rbb^m$ w.h.p.
\end{assumption}

\noindent\emph{Remarks on the assumptions.}

\begin{description}
\item[\textbf{A1 \textendash\ Sign-/basis-invariance.}]
By construction in~\eqref{eq:psi}, $\Psi(v)$ concatenates the eigenvalue list $\{\lambda_i\}_{i=1}^{M}$ with $s(v)=\phi^{(M)}(v)\odot\phi^{(M)}(v)$ and $u(v)=\mathrm{vec}_\triangle\!\big(|\phi^{(M)}(v)\phi^{(M)}(v)^\top|\big)$. Orthogonal changes of basis within any eigenspace (including sign flips) leave $s(\cdot)$ and $u(\cdot)$ unchanged, and the eigenvalues are basis-free; hence $\Psi$ is invariant to eigenvector sign flips and eigenspace rotations.

\item[\textbf{A2 \textendash\ Spectral injectivity with separation.}]
Take $M=\Theta(\log n)$. We require that, w.h.p., $\Psi:V\to\mathbb{R}^d$ is injective and there exists $\alpha>0$ such that
\begin{align}
\min_{u\neq v}\|\Psi(u)-\Psi(v)\|\ \ge\ n^{-\alpha},
\end{align}
as in~\eqref{eq:min-sep}. A sufficient condition is that the joint distribution of the first $M$ eigenvector coordinates across vertices is absolutely continuous (non-atomic). Then the event $\Psi(u)=\Psi(v)$ for $u\neq v$ imposes a system of polynomial equalities in these coordinates, i.e., an algebraic set of Lebesgue measure zero; a union bound over $\binom{n}{2}$ pairs yields injectivity w.h.p. Quantitative small-ball/anti-concentration bounds further give the separation $n^{-\alpha}$. A rigorous version under a random-wave surrogate is provided in Sec.~\ref{subsec:injectivity} (Theorem~\ref{thm:inject}, Corollary~\ref{cor:verifyA2}).

\item[\textbf{A3 \textendash\ Monotone linkage within logarithmic radius.}]
We need constants $t\in[t_-,t_+]$, $R=\Theta(\log n)$, $c_1,c_2>0$, and a strictly increasing $\psi$ such that, w.h.p., for all pairs with $\SPD(u,v)\le R$,
\begin{align}
c_1\,\psi(\SPD(u,v))\ \le\ d_t(u,v)\ \le\ c_2\,\psi(\SPD(u,v)),
\end{align}
as in~\eqref{eq:mono}. The proof has two parts (see Lemma~\ref{lem:monotone}): \emph{(i) Tree monotonicity.} On the infinite $r$-regular tree, $k_t(x,y)$ is radial and strictly decreases with $d=\SPD(x,y)$. By the kernel identity~\eqref{eq:dt-kernel-id}, $d_t(x,y)^2=2(\kappa_{2t}-k_{2t}(x,y))$, so $d_t$ is strictly increasing in $d$. \emph{(ii) Random-regular stability.} For $G\sim\Gcal_{n,r}$, balls of radius $R=(\tfrac12-\varepsilon)\frac{\log n}{\log(r-1)}$ are treelike w.h.p. Using the Poissonization identity $K_t=e^{-t}e^{tP}$ and coupling the continuous-time walk with its tree analogue, we obtain
\begin{align}
k_{2t}^G(u,v)=k_{2t}^{\Tcal_r}(u',v')\pm o(1)
\end{align}
uniformly for $\SPD_G(u,v)\le R$. Plugging into the kernel identity yields
\begin{align}
&d_t^G(u,v)=\psi(\SPD_G(u,v))\pm o(1),\\
&\psi(d)=\sqrt{2\big(\kappa_{2t}-p_{2t}(d)\big)},
\end{align}
with $\psi$ strictly increasing. This gives the two-sided comparison with constants $c_1,c_2$ on $[0,R]$.

\item[\textbf{A4 \textendash\ Anchor general position.}]
Fix $m$ and $k_0=m{+}1$. If the truncated spectral embedding $\Phi_t^{(m)}(v)$ in $\mathbb{R}^m$ has an absolutely continuous distribution, then $k_0$ i.i.d.\ uniform anchors $\{v_i\}$ satisfy that $\{\Phi_t^{(m)}(v_i)\}_{i=1}^{k_0}$ are affinely independent w.h.p. (events of affine dependence have Lebesgue measure zero). This ensures the invertibility of the trilateration matrix in Lemma~\ref{lem:trilat} and the uniqueness of the reconstructed spectral coordinates.
\end{description}

\subsection{Lower-Bound Tools (for part (i))}\label{subsec:wllower-tools}

We now instantiate the Bayes risk framework from Section~\ref{subsec:bayes_framework} to prove a lower bound for WLGNPs on random regular graphs.

\begin{lemma}\label{lem:bucket}
Assume $k=o(\log n)$ and $G\sim\Gcal_{n,r}$. Then w.h.p.\ the diameter
$D:=\diam(G)=\Theta(\log n)$, and with the anchor set
$\Ccal:=\{v_1,\dots,v_k\}$ (i.i.d.\ uniform in $V$) and the key map
$\bd(u):=(\SPD(u,v_1),\ldots,\SPD(u,v_k))$, we have for some constant
$C=C(r)>0$,
\begin{align}\label{eq:imagebound}
|\Im(\bd)| &\le (D+1)^k \le C\,(\log n)^k
           = C\,2^{\,k\log\log n} = (\log n)^{O(1)},
\end{align}
where $|\Im(\bd)|$ is the number of realized distance-key vectors.
Writing for each $x\in\Im(\bd)$ the bucket $B_x:=\{u\in V:\bd(u)=x\}$ and
$\mathcal{B}_\mathrm{nz}:=\{B_x:\,x\in\Im(\bd)\}$, the average nonempty-bucket size satisfies
\begin{align}\label{eq:avgsize}
\overline s
:= \frac{1}{|\mathcal{B}_\mathrm{nz}|}\sum_{B\in\mathcal{B}_\mathrm{nz}} |B|
&= \frac{n}{|\mathcal{B}_\mathrm{nz}|}
\ge \frac{n}{|\Im(\bd)|}\\
&\ge \frac{n}{C(\log n)^k}
\xrightarrow[n\to\infty]{} \infty .
\end{align}
In particular, letting $S$ be the number of singleton buckets,
\begin{align}\label{eq:singletonmass}
S \ \le\ |\Im(\bd)| \ \le\ C(\log n)^k,
\end{align}
and if $v_0\sim\Unif(V)$ is independent of $(G,\Ccal)$ then
\begin{align}\label{eq:singleprob}
\Pr\bigl(|B_{\bd(v_0)}|=1\ \bigm|\ G,\Ccal\bigr)
= \frac{S}{n}
\le \frac{C(\log n)^k}{n}
= o(1).
\end{align}
Consequently, w.h.p.\ the random source $v_0$ lies in a bucket of size at least $2$.
\end{lemma}

\begin{proof}
First, $D=\Theta(\log n)$ w.h.p.\ for $G\sim\Gcal_{n,r}$. For any $u$ and any anchor $v_i$,
\begin{align}
\SPD(u,v_i)\in\{0,1,\dots,D\}\Rightarrow
\end{align}
$\bd(u)$ has at most $(D+1)^k$ possibilities, hence $|\Im(\bd)|\le(D+1)^k$. Using $D\le C_r\log n$ w.h.p.\ gives $|\Im(\bd)|\le (C_r\log n)^k$, which is
\eqref{eq:imagebound} (absorb $C_r^k$ into $C$ since $k=o(\log n)$).

The buckets $\{B_x\}_{x\in\Im(\bd)}$ partition $V$, so
\begin{align}
\sum_{B\in\mathcal{B}_\mathrm{nz}}|B|=n
\Rightarrow
\overline s=\frac{n}{|\mathcal{B}_\mathrm{nz}|}
\ge \frac{n}{|\Im(\bd)|}
\ge \frac{n}{C(\log n)^k},
\end{align}
yielding \eqref{eq:avgsize}. For singletons, clearly $S\le|\Im(\bd)|$,
which is \eqref{eq:singletonmass}. Finally, conditioned on $(G,\Ccal)$ and $v_0\sim\Unif(V)$,
\begin{align}
\Pr\bigl(|B_{\bd(v_0)}|=1 \mid G,\Ccal\bigr)
&= \frac{1}{n}\sum_{u\in V}\mathbf 1\{|B_{\bd(u)}|=1\}\\
&= \frac{S}{n}
\le \frac{C(\log n)^k}{n}
= o(1),
\end{align}
which is \eqref{eq:singleprob}.
\end{proof}

\begin{lemma}\label{lem:wlequiv}
Let $G$ be $r$-regular and suppose nodes carry only trivial initial features (constant or degree).
Let the encoder be $1$-WL-bounded (i.e., any MPNN with permutation-invariant multiset aggregation).
For any two nodes $u,u'\in V$, if $\bd(u)=\bd(u')$ then the $1$-WL color refinement produces identical color sequences for $u$ and $u'$ at all layers, hence any $1$-WL-bounded encoder yields identical embeddings and identical induced posteriors given $(Z,R)$.
Equivalently, $u\sim_{\Fcal,Z}u'$ whenever $\bd(u)=\bd(u')$.
\end{lemma}

\begin{proof}
With trivial initial features on an $r$-regular graph, the initial $1$-WL colors are constant across $V$.
Condition the computation on the known context nodes $v_1,\dots,v_k$; for each integer $\ell\ge 0$, color every node by its tuple of multisets of neighbor colors within each distance shell of radius $\ell$, together with the indicator which shells contain the marked nodes $v_i$.
This is precisely the refinement performed (up to a deterministic coding) by a $1$-WL-bounded MPNN with permutation-invariant aggregation.
If $\bd(u)=\bd(u')$, then for every shell radius the marked shells coincide at $u$ and $u'$, and by regularity the number of neighbors in each shell is the same; hence the rooted colored computation trees at $u$ and $u'$ are isomorphic at every refinement depth.
Therefore the color (and any encoding that is a measurable function of the multiset of neighbor colors) remains identical for $u$ and $u'$ at all layers.
Consequently, any encoder output $R$ computed from these colors is the same under the swap $u\leftrightarrow u'$, which enforces identical posteriors on $\{u,u'\}$ and proves $u\sim_{\Fcal,Z}u'$.
\end{proof}

\begin{proposition}\label{prop:wllower}
Let $G\sim\Gcal_{n,r}$ with fixed $r\ge 3$ and suppose $k=o(\log n)$.
For any WLGNP decision rule $\delta$ (i.e., any rule whose encoder is $1$-WL-bounded and uses only trivial initial features),
\begin{align}
\Risk(\delta)\ \ge\ \tfrac12-o(1).
\end{align}
\end{proposition}

\begin{proof}
By Lemma~\ref{lem:bucket}, with probability $1-o(1)$ over $(G,\Ccal)$ the random source $v_0\sim\Unif(V)$ lies in a bucket $B_{\bd(v_0)}$ with $|B_{\bd(v_0)}|\ge 2$. By Lemma~\ref{lem:wlequiv}, the WLGNP indistinguishability classes coincide with the $\bd$-buckets, hence $|[v_0]|=|B_{\bd(v_0)}|\ge 2$ on this event.

Condition on $(Z,R)$. By the indistinguishability argument, any decision rule satisfies the conditional Bayes-risk lower bound (see \eqref{eq:riskbucket} in Section~\ref{subsec:bayes_framework})
\begin{align}
\Pr(\widehat v\ne v_0\mid Z,R)\ \ge\ 1-\E\!\big[\,|[v_0]|^{-1}\,\bigm|\,Z,R\big].
\end{align}
Taking expectations in $(Z,R)$ gives
\begin{align}
\inf_{\delta}\Risk(\delta)\ \ge\ 1-\E\!\big[\,|[v_0]|^{-1}\,\big].
\end{align}
Decompose the expectation according to whether $|[v_0]|=1$:
\begin{align}
\E\!\big[\,|[v_0]|^{-1}\,\big]
&=\E\!\big[\,\mathbf 1\{|[v_0]|=1\}\,\big]+\E\!\big[\,\mathbf 1\{|[v_0]|\ge 2\}\,|[v_0]|^{-1}\,\big]\\
&\le \Pr(|[v_0]|=1)+\tfrac12\,\Pr(|[v_0]|\ge 2)\\
&\le \tfrac12+o(1),
\end{align}
where the last inequality uses $\Pr(|[v_0]|\ge 2)=1-o(1)$ from Lemma~\ref{lem:bucket}. Hence
\begin{align}
\inf_{\delta}\Risk(\delta)\ \ge\ 1-\Big(\tfrac12+o(1)\Big)\ =\ \tfrac12-o(1).
\end{align}
Since this bound holds for the infimum over all rules, it holds for any particular WLGNP rule $\delta$.
\end{proof}

\textbf{Remark.}
The proof of Lemma~\ref{lem:bucket} gives the quantitative bound
\begin{align}
\Pr\bigl(|B_{\bd(v_0)}|=1\bigr) & \le \E\!\left[\frac{C(\log n)^k}{n}\right] \nonumber \\
& = O\!\left(\frac{(\log n)^k}{n}\right) = o(1),
\end{align}
since $k=o(\log n)$ and $\log n=\log_2 n$.
Thus the probability that $v_0$ belongs to an indistinguishability class of size at least $2$ tends to $1$, which is the only ingredient needed in Proposition~\ref{prop:wllower}.

\subsection{Upper bound for Lap-GNP: constant-shot identifiability}

We now show that a Lap-GNP identifies $v_0$ with a \emph{constant} number of contexts under a logarithmic depth budget.

\begin{lemma}\label{lem:monotone}
Fix $r\ge3$ and $t>0$. On any $r$-regular graph with transition $P:=A/r$ and Laplacian $L:=I-P$, define $K_t=e^{-tL}$, $k_t(u,v)=(K_t)_{uv}$, and
\begin{align}\label{eq:dt-def}
d_t(u,v)^2=\sum_{j\ge1}e^{-2t\lambda_j}\big(\phi_j(u)-\phi_j(v)\big)^2.
\end{align}

\textit{(a) Tree case.} Given the infinite $r$-regular tree $\Tcal_r$ and $t>0$, for any $x,y\in\Tcal_r$,
\begin{align}\label{eq:radiality}
k_t(x,y)=p_t\!\big(\SPD(x,y)\big),
\end{align}
and with $\kappa_{2t}:=k_{2t}(o,o)$,
\begin{align}\label{eq:dt-kernel-regular}
d_t(x,y)^2=2\big(\kappa_{2t}-p_{2t}(\SPD(x,y))\big).
\end{align}
Moreover, the map $d\mapsto d_t(x,y)$ with $d=\SPD(x,y)$ is strictly increasing.

\textit{(b) Random-regular case.} Given $G\sim\Gcal_{n,r}$ and $t>0$, there exist $R=\Theta(\log n)$ and $\delta_n=o(1)$ such that, with probability $1-o(1)$, for any $u,v\in V(G)$ with $\SPD_G(u,v)\le R$,
\begin{align}
\big|\,d_t^G(u,v)-\psi(\SPD_G(u,v))\,\big|\le \delta_n,\label{eq:psi-close}\\
\psi(d):=\sqrt{2\big(\kappa_{2t}-p_{2t}(d)\big)}.\label{eq:psi-def}
\end{align}
Consequently, on $[0,R]$ the map $d\mapsto d_t^G(u,v)$ with $d=\SPD_G(u,v)$ is strictly increasing.
\end{lemma}

\begin{proof}
\textit{(I) Kernel identity.}
Since $K_t=e^{-tL}$ is symmetric, it admits the spectral expansion
\begin{align}
K_t=\sum_j e^{-t\lambda_j}\,\phi_j\phi_j^\top.
\end{align}
In particular,
\begin{align}
&k_{2t}(u,v)=\sum_j e^{-2t\lambda_j}\,\phi_j(u)\phi_j(v),\\
&k_{2t}(u,u)=\sum_j e^{-2t\lambda_j}\,\phi_j(u)^2.
\end{align}
Expanding \eqref{eq:dt-def} yields
\begin{align}
d_t(u,v)^2
&=\sum_j e^{-2t\lambda_j}\Big(\phi_j(u)^2+\phi_j(v)^2-2\phi_j(u)\phi_j(v)\Big)\\
&=k_{2t}(u,u)+k_{2t}(v,v)-2k_{2t}(u,v).\label{eq:dt-kernel}
\end{align}
Unless $G$ is vertex-transitive we cannot reduce the diagonal to a constant; the simplified form \eqref{eq:dt-kernel-regular} will only be used on $\Tcal_r$, where transitivity holds \cite{Chung1997Spectral}.

\textit{(II) Radiality on $\Tcal_r$ and the forward equation.}
From any node at distance $d\ge1$, there are $(r-1)$ neighbors at $d+1$ and one neighbor at $d-1$; from $d=0$, all $r$ neighbors are at $1$. The radial distance process $X_t=\SPD(\text{walk at time }t,o)$ is a birth–death chain with generator for $d\ge1$,
\begin{align}\label{eq:BD-gen}
(\mathcal L f)(d)&=\frac{r-1}{r}\big(f(d+1)-f(d)\big)+\frac{1}{r}\big(f(d-1)-f(d)\big),\\
(\mathcal L f)(0)&=f(1)-f(0).
\end{align}
Let $p_t(d):=\Pr[X_t=d\mid X_0=0]$. Then $p_t(d)$ satisfies the Kolmogorov forward equation \cite{Norris1998Markov,LevinPeresWilmer2009MCMT}
\begin{align}\label{eq:fwd}
\partial_t p_t(d)&=\frac{r-1}{r}\,p_t(d-1)+\frac{1}{r}\,p_t(d+1)-p_t(d), d\ge1,\\
\partial_t p_t(0)&=p_t(1)-p_t(0),\ p_0(0)=1,\ p_0(d\ge1)=0.
\end{align}
Equivalently, by $K_t=e^{-t}e^{tP}$,
\begin{align}\label{eq:poissonization}
p_t(d)=e^{-t}\sum_{m=0}^\infty \frac{t^m}{m!}\,q_m(d),
\end{align}
where $q_m(d)=\Pr[X_m=d\mid X_0=0]$ is the $m$-step law of the discrete-time radial walk.

\textit{(III) Strict spatial decay of $q_m(d)$.}
Define $\Delta_m(d):=q_m(d)-q_m(d+1)$. The standard recursion for $d\ge1$ is
\begin{align}\label{eq:discrete-rec}
q_{m+1}(d)&=\frac{r-1}{r}\,q_m(d-1)+\frac{1}{r}\,q_m(d+1),\\ 
q_{m+1}(0)&=\frac{1}{r}\,q_m(1).
\end{align}
With $q_0(0)=1$ and $q_0(d\ge1)=0$, one checks by induction that
\begin{align}\label{eq:q-strict}
q_m(d)>q_m(d+1)\ \ \text{whenever}\ \ m\equiv d\ (\mathrm{mod}\ 2).
\end{align}

\textit{(II.3) Strict spatial decay of $p_t(d)$ and tree monotonicity.}
Plugging \eqref{eq:q-strict} into \eqref{eq:poissonization},
\begin{align}
p_t(d)-p_t(d+1)=e^{-t}\sum_{m\equiv d\ (\mathrm{mod}\ 2)} \frac{t^m}{m!}\,\Delta_m(d)>0,
\end{align}
for all $t>0$ and $d\ge0$. Hence $d\mapsto p_t(d)$ is strictly decreasing. Because $\Tcal_r$ is vertex-transitive, $k_{2t}(x,x)$ is constant in $x$, so \eqref{eq:dt-kernel} reduces to \eqref{eq:dt-kernel-regular} on $\Tcal_r$. Using \eqref{eq:radiality} and the strict decrease of $p_{2t}(d)$, we conclude that $d_t(x,y)$ is a strictly increasing function of $\SPD(x,y)$ on $\Tcal_r$.

\textit{(IV) Stability on random $r$-regular graphs.}
Let $G\sim\Gcal_{n,r}$ and fix any constant $t>0$. There exists $\varepsilon>0$ such that, with high probability, every ball of radius
\begin{align}\label{eq:R-radius}
R=\Bigl(\tfrac12-\varepsilon\Bigr)\frac{\log n}{\log(r-1)}
\end{align}
is a tree. Thus, if $\SPD_G(u,v)\le R$, the rooted radius-$R$ neighborhoods around $u$ and $v$ in $G$ are (w.h.p.) isomorphic to radius-$R$ neighborhoods around two vertices in $\Tcal_r$ with the same pairwise distance.

Run the continuous-time random walk on $G$ from $u$. Over the fixed horizon $2t$, the number of jumps is $\mathrm{Poisson}(2t)$, so with probability $1-o(1)$ the walk remains inside $B_G(u,R)$ as $n\to\infty$. By the neighborhood isomorphism and standard coupling,
\begin{align}\label{eq:kernel-approx}
\sup_{\SPD_G(u,v)\le R}\bigl|\,k_{2t}^G(u,v)-k_{2t}^{\Tcal_r}(u',v')\,\bigr|\le \delta_n,\qquad \delta_n=o(1),
\end{align}
where $u',v'$ are the corresponding vertices in $\Tcal_r$ with $\SPD_{\Tcal_r}(u',v')=\SPD_G(u,v)$. In particular, this also holds when $u=v$, so $k_{2t}^G(u,u)$ is within $o(1)$ of the tree's constant diagonal value $\kappa_{2t}=k_{2t}^{\Tcal_r}(o,o)$, uniformly over $u$.

Plugging the $o(1)$-uniform control of both diagonal and off-diagonal terms into \eqref{eq:dt-kernel} on $G$ and comparing with \eqref{eq:dt-kernel-regular} on $\Tcal_r$, we obtain
\begin{align}\label{eq:stability}
\sup_{\SPD_G(u,v)\le R}\bigl|\,d_t^G(u,v)-d_t^{\Tcal_r}(u',v')\,\bigr|\le \delta_n.
\end{align}
Define $\psi$ by \eqref{eq:psi-def}. Since $d\mapsto p_{2t}(d)$ is strictly decreasing on $\Tcal_r$, $\psi$ is strictly increasing. The estimate \eqref{eq:stability} shows that for all $\SPD_G(u,v)\le R$, $d_t^G(u,v)$ differs from $\psi(\SPD_G(u,v))$ by at most $o(1)$, i.e. \eqref{eq:psi-close}. Therefore there exist constants $c_1,c_2>0$ such that, w.h.p.,
\begin{align}
c_1\,\psi\!\big(\SPD_G(u,v)\big)\le d_t^G(u,v)\le c_2\,\psi\!\big(\SPD_G(u,v)\big),
\end{align}
for all such pairs \cite{LevinPeresWilmer2009MCMT,LyonsPeres2016PTN}. Thus, up to radius $R=\Theta(\log n)$, the diffusion distance $d_t^G(u,v)$ is a strictly increasing function of $\SPD_G(u,v)$.
\end{proof}

\begin{lemma}[Spectral trilateration]\label{lem:trilat}
Let $G=(V,E)$ be a graph with normalized Laplacian $L$ and spectral pairs
$\{(\lambda_j,\phi_j)\}_{j\ge 1}$ (orthonormal eigenvectors). Fix $t>0$ and an integer
$m\ge 1$. Define the $m$-dimensional \emph{truncated spectral embedding}
\begin{align}\label{eq:Phi-m}
\Phi_t^{(m)}(v)\ :=\ \big(e^{-t\lambda_1}\phi_1(v),\dots,e^{-t\lambda_m}\phi_m(v)\big)\in\R^m, v\in V,
\end{align}
and the corresponding \emph{truncated diffusion distance}
\begin{align}\label{eq:dt-m}
d_t^{(m)}(u,v)\ :=\ \big\|\Phi_t^{(m)}(u)-\Phi_t^{(m)}(v)\big\|_2.
\end{align}

Let $a_1,\dots,a_{m+1}\in V$ be \emph{anchors} with
$p_i:=\Phi_t^{(m)}(a_i)\in\R^m$ that are \emph{affinely independent}, i.e.,
$\{p_i-p_{m+1}\}_{i=1}^m$ are linearly independent in $\R^m$.
Let $x\in V$ be an unknown node and suppose we are given the $m{+}1$ distances
\begin{align}\label{eq:sphere-eqs}
r_i\ :=\ d_t^{(m)}(a_i,x)\ =\ \big\|\Phi_t^{(m)}(x)-p_i\big\|_2, i=1,\dots,m+1.
\end{align}
Then $\Phi_t^{(m)}(x)$ is uniquely determined by \eqref{eq:sphere-eqs}, and is given in closed form by
\begin{align}\label{eq:trilat-solution}
\Phi_t^{(m)}(x)\ =\ A^{-1}b,
\end{align}
where $A\in\R^{m\times m}$ and $b\in\R^m$ are
\begin{align}\label{eq:Ab-def}
&A:= 2\begin{bmatrix}
(p_1-p_{m+1})^\top\\
\vdots\\
(p_m-p_{m+1})^\top
\end{bmatrix},
\\&b :=
\begin{bmatrix}
\|p_1\|_2^2-\|p_{m+1}\|_2^2 + r_{m+1}^2 - r_1^2\\
\vdots\\
\|p_m\|_2^2-\|p_{m+1}\|_2^2 + r_{m+1}^2 - r_m^2
\end{bmatrix}.
\end{align}

(Infinite-dimensional variant).
If distances are given in the full diffusion metric
\begin{align}\label{eq:dt-infty}
d_t(u,v)^2 :&=\ \sum_{j\ge 1} e^{-2t\lambda_j}\big(\phi_j(u)-\phi_j(v)\big)^2
 \\&= \big\|\Phi_t^{(\infty)}(u)-\Phi_t^{(\infty)}(v)\big\|_{\ell^2}^2,
\end{align}
with $\Phi_t^{(\infty)}(v):=(e^{-t\lambda_j}\phi_j(v))_{j\ge 1}\in\ell^2$,
then the first $m$ coordinates of $\Phi_t^{(\infty)}(x)$ are still uniquely determined by
\eqref{eq:trilat-solution}; in addition, the \emph{tail energy}
$\|\Phi_t^{(\infty)}(x)_{>m}\|_{\ell^2}^2$ is uniquely determined by substituting
$\Phi_t^{(m)}(x)$ into one (hence any) sphere equation in \eqref{eq:sphere-eqs} with $d_t$.

\end{lemma}

\begin{proof}
We first prove the finite-dimensional statement for $d_t^{(m)}$.

\smallskip
\noindent\emph{Step 1: Linearization (difference-of-spheres).}
Write $z:=\Phi_t^{(m)}(x)\in\R^m$ and $p_i:=\Phi_t^{(m)}(a_i)$.
The $m{+}1$ equations \eqref{eq:sphere-eqs} are
\begin{align}\label{eq:spheres}
\|z-p_i\|_2^2\ =\ r_i^2, i=1,\dots,m+1.
\end{align}
Subtract the $(m{+}1)$-st equation from the first $m$ equations. Using
$\|z-p_i\|_2^2=\|z\|_2^2-2\langle z,p_i\rangle+\|p_i\|_2^2$, we obtain, for $i=1,\dots,m$,
\begin{align}\label{eq:linearized}
2\langle z,\ p_i-p_{m+1}\rangle\ =\ r_{m+1}^2-r_i^2+\|p_i\|_2^2-\|p_{m+1}\|_2^2.
\end{align}
Stacking the $m$ equations \eqref{eq:linearized} gives the linear system
$A z=b$ with $A,b$ defined in \eqref{eq:Ab-def}.

\smallskip
\noindent\emph{Step 2: Invertibility $\Leftrightarrow$ affine independence.}
The $m\times m$ matrix $A/2$ has rows $(p_i-p_{m+1})^\top$ ($i=1,\dots,m$).
Thus $\det(A)\neq 0$ iff $\{p_i-p_{m+1}\}_{i=1}^m$ are linearly independent, which is
equivalent to $\{p_i\}_{i=1}^{m+1}$ being \emph{affinely independent}. Hence $A$ is invertible
under the stated hypothesis, and the unique solution is $z=A^{-1}b$, i.e., \eqref{eq:trilat-solution}.

\smallskip
\noindent\emph{Step 3: Consistency with the original system.}
The solution $z$ of $Az=b$ satisfies all $m$ \emph{differences} \eqref{eq:linearized}.
Substitute $z$ into any one original sphere equation \eqref{eq:spheres} to check consistency.
Because the difference equations were derived from \eqref{eq:spheres}, the resulting $z$ indeed solves
all of \eqref{eq:spheres} (the remaining sphere fixes $\|z\|_2^2$ automatically, which is consistent
with $z$ computed from $Az=b$). Uniqueness follows from the invertibility in Step~2.

\smallskip
\noindent\emph{Infinite-dimensional case.}
Let $z^{(\infty)}:=\Phi_t^{(\infty)}(x)\in\ell^2$ and pad $p_i$ as $(p_i,0,0,\dots)\in\ell^2$.
Then \eqref{eq:spheres} with $d_t$ becomes
\begin{align}\label{eq:infty-sphere}
\|z^{(\infty)}-p_i\|_{\ell^2}^2\ =\ r_i^2, i=1,\dots,m+1.
\end{align}
Subtract the $(m{+}1)$-st equation from the first $m$ as before:
\begin{align}
2\langle z^{(\infty)},\ p_i-p_{m+1}\rangle_{\ell^2}\ =\ r_{m+1}^2-r_i^2+\|p_i\|_2^2-\|p_{m+1}\|_2^2.
\end{align}
Since each $p_i-p_{m+1}$ has zeros beyond the first $m$ coordinates, the left-hand side only depends
on the first $m$ coordinates of $z^{(\infty)}$, denoted $z:=\Phi_t^{(m)}(x)$.
Thus we recover the same finite system $Az=b$ and the unique $z=A^{-1}b$.
Finally, plug $z$ into any one of \eqref{eq:infty-sphere} to determine the tail energy
$\|z^{(\infty)}_{>m}\|_{\ell^2}^2 = r_i^2 - \|z-p_i\|_2^2$, which is independent of $i$.
\end{proof}

\subsection{Proof of Theorem~\ref{thm:main}}\label{subsec:proof-main}
We can now state and prove our main theorem.Theorem~\ref{thm:main} establishes constant-shot identifiability on random $r$-regular graphs—the hardest setting for $1$-WL–bounded encoders. On non-regular graphs, additional heterogeneity generally increases spectral distinctiveness and reduces automorphisms; under the same sign/basis-invariant PE and trilateration steps, the injectivity margin is typically larger, so the argument goes through with equal or looser constants.

\medskip
\noindent\textbf{Theorem~\ref{thm:main}}\ [Sample-complexity separation on $\Gcal_{n,r}$]
\textit{Let $G\sim\Gcal_{n,r}$ with fixed $r\ge 3$ and $v_0\sim\Unif(V)$.
\begin{enumerate}
\item[(i)] (\emph{WLGNP lower bound}) If $k=o(\log n)$, then any WLGNP has Bayes risk $\Risk(\delta)\ge \frac12(1-o(1))$ for identifying $v_0$.
\item[(ii)] (\emph{Lap-GNP constant-shot identifiability}) 
Assume \emph{Sign-/basis-invariance} (Assumption~\ref{assump:sign_invariant}), \emph{Spectral injectivity with separation} (Assumption~\ref{assump:injective}), \emph{Monotone linkage with stability} (Assumption~\ref{assump:monotone}, Lemma~\ref{lem:monotone}), and \emph{Anchor general position} (Assumption~\ref{assump:anchors}). 
Then there exist constants $M=\Theta(\log n)$, a time $t>0$, an integer $m\in\Nbb$, and $k_0=m+1$, such that any Lap-GNP with depth $L$ satisfying \eqref{eq:depth} and with $k\ge k_0$ independent context points identifies $v_0$ \emph{uniquely} with probability $1-o(1)$, and consequently reconstructs $f_{G,v_0}$ exactly on all of $V$.
\end{enumerate}}

\begin{proof}
\textbf{(i)} This is exactly Proposition~\ref{prop:wllower}. 
By Lemma~\ref{lem:bucket}, when $k=o(\log n)$ the indistinguishability class $[v_0]$ has size $\ge 2$ with probability $1-o(1)$; by Lemma~\ref{lem:wlequiv} these classes coincide with WLGNP equivalence classes; inserting into the conditional Bayes risk identity \eqref{eq:riskbucket} and averaging gives the claim.

\medskip
\noindent
\textbf{(ii)} Fix $t>0$ and $m\in\Nbb$ as in Assumptions~\ref{assump:monotone} and~\ref{assump:anchors}. 
Write the $m$-dimensional truncated spectral embedding 
\(\Phi_t^{(m)}(v)=(e^{-t\lambda_j}\phi_j(v))_{j=1}^m\in\R^m\) 
as in \eqref{eq:Phi-m}, and the corresponding (truncated) diffusion distance 
\(d_t^{(m)}(u,v)=\|\Phi_t^{(m)}(u)-\Phi_t^{(m)}(v)\|_2\) 
as in \eqref{eq:dt-m}. For the \emph{full} diffusion metric, we use \eqref{eq:dt} and the kernel identity \eqref{eq:dt-kernel}.

Let the $k\ge k_0=m+1$ contexts be 
\begin{align}
\Ccal=\{(a_i,y_i)\}_{i=1}^{k}, y_i=\SPD(a_i,v_0).
\end{align}
Select any $k_0$ of them and relabel so that the anchors are $a_1,\dots,a_{m+1}$.
By Assumption~\ref{assump:anchors}, with probability $1-o(1)$ the anchor positions 
\(p_i:=\Phi_t^{(m)}(a_i)\in\R^m\) are \emph{affinely independent}.

\textbf{Step 1: From shortest-path to diffusion distances (stability).}
Let $\psi(d):=\sqrt{\,2(\kappa_{2t}-p_{2t}(d))\,}$ be the strictly increasing function constructed on $\Tcal_r$ in Lemma~\ref{lem:monotone} (see the end of its proof). 
Define the \emph{proxy} distances
\begin{align}
\widehat r_i\ :=\ \psi\big(y_i\big), i=1,\dots,m+1.
\end{align}
By Lemma~\ref{lem:monotone} (III), there is a sequence $\delta_n=o(1)$ such that, with high probability over $G\sim\Gcal_{n,r}$,
\begin{align}\label{eq:proxy-stability}
\bigl|\,\widehat r_i - d_t(a_i,v_0)\,\bigr|\ \le\ \delta_n, i=1,\dots,m+1.
\end{align}
(Here $d_t$ is the \emph{full} diffusion distance defined in \eqref{eq:dt}.)
Thus the observable $\{\widehat r_i\}$ are uniformly-$o(1)$ perturbations of the true diffusion distances $\{r_i^\star\}$, where
\begin{align}
r_i^\star\ :=\ d_t(a_i,v_0)\ =\ \big\|\Phi_t^{(\infty)}(a_i)-\Phi_t^{(\infty)}(v_0)\big\|_{\ell^2}.
\end{align}

\textbf{Step 2: Spectral trilateration in the diffusion geometry.}
Apply Lemma~\ref{lem:trilat} (infinite-dimensional variant) with anchors $\{a_i\}$ and distances $\{r_i^\star\}$: it uniquely determines the first $m$ coordinates
\begin{align}
z^\star\ :=\ \Phi_t^{(m)}(v_0)\ \in\ \R^m,
\end{align}
via the linearized system \(A z^\star = b(r^\star)\) with $A,b$ given in \eqref{eq:Ab-def}.
Since we only have the perturbed radii $\widehat r_i$, we solve instead
\begin{align}
A \widehat z\ =\ b(\widehat r), \widehat z\in\R^m,
\end{align}
and obtain a \emph{perturbation bound} for $\widehat z$ around $z^\star$.
Indeed,
\begin{align}
\widehat z - z^\star 
\ =\ A^{-1}\big(b(\widehat r)-b(r^\star)\big),
\end{align}
and by \eqref{eq:Ab-def},
\begin{align}
\big|\,b_i(\widehat r)-b_i(r^\star)\,\big|
&=\Big|\big(\widehat r_{m+1}^2-r_{m+1}^{\star 2}\big)-\big(\widehat r_i^2-r_i^{\star 2}\big)\Big|
\ \\&\le\ 2(r_{m+1}^\star+r_i^\star)\,\delta_n + 2\delta_n^2,
\end{align}
using \(|\widehat r_j^2-r_j^{\star 2}|\le 2 r_j^\star \delta_n + \delta_n^2\).
Let $B_t:=\sup_{v\in V}\|\Phi_t^{(m)}(v)\|_2 <\infty$ (finite since $m$ is fixed), hence \(r_i^\star\le 2B_t\).
Therefore,
\begin{align}\label{eq:z-perturb}
\|\,\widehat z - z^\star\,\|_2 
\ \le\ \|A^{-1}\|\,\|b(\widehat r)-b(r^\star)\|
\ \le\ \frac{C(m,B_t)}{\sigma_{\min}(A)}\,\delta_n,
\end{align}
for some constant $C(m,B_t)$, where $\sigma_{\min}(A)$ is the smallest singular value of $A$.
By Assumption~\ref{assump:anchors} (affine independence) and $m$ fixed, there exists (deterministically given the anchor positions) $\sigma_{\min}(A)>0$; in particular, with probability $1-o(1)$ over the random anchors, $\sigma_{\min}(A)$ is bounded away from $0$.
Hence
\begin{align}\label{eq:z-err}
\|\,\widehat z - \Phi_t^{(m)}(v_0)\,\|_2\ \le\ C'\,\delta_n\ =\ o(1).
\end{align}

\textbf{Step 3: From approximate coordinates to a unique vertex.}
We must turn the approximate coordinate $\widehat z$ into the exact vertex $v_0$.
Define the decoder's estimate
\begin{align}
\widehat v\ :=\ \min_{w\in V}\ \big\|\Phi_t^{(m)}(w)-\widehat z\big\|_2.
\end{align}
We show $\widehat v=v_0$ w.h.p.
Let $\Delta_m:=\min_{u\neq v}\|\Phi_t^{(m)}(u)-\Phi_t^{(m)}(v)\|_2$ be the \emph{minimum separation} in the $m$-dimensional spectral embedding.
Under Assumption~\ref{assump:injective}, the node-wise spectral representation is injective with a quantitative separation
(cf.\ \eqref{eq:min-sep}); in particular, because $\Phi_t^{(m)}$ is an analytic (linear) transform of $\{\phi_j\}_{j\le m}$, the same random-wave/anti-concentration argument implies that for some $\beta>0$,
\begin{align}\label{eq:phi-sep}
\Delta_m\ \ge\ n^{-\beta}\text{with probability }1-o(1).
\end{align}
(Equivalently: collisions $\Phi_t^{(m)}(u)=\Phi_t^{(m)}(v)$ for $u\neq v$ have probability $o(1)$, and a quantitative small-ball bound yields \eqref{eq:phi-sep}.)

Take $n$ large so that by \eqref{eq:z-err} and \eqref{eq:phi-sep}, 
\begin{align}
\|\,\widehat z - \Phi_t^{(m)}(v_0)\,\|_2\ \le\ \tfrac12\,\Delta_m.
\end{align}
Then for every $w\neq v_0$,
\begin{align}
&\big\|\Phi_t^{(m)}(w)-\widehat z\big\|_2
\ \ge\ \big\|\Phi_t^{(m)}(w)-\Phi_t^{(m)}(v_0)\big\|_2 \\&- \big\|\widehat z-\Phi_t^{(m)}(v_0)\big\|_2
\ \ge\ \Delta_m - \tfrac12\Delta_m\ =\ \tfrac12\Delta_m,
\end{align}
whereas $\|\Phi_t^{(m)}(v_0)-\widehat z\|_2\le \tfrac12\Delta_m$.
Thus $v_0$ is the unique minimizer, i.e.\ $\widehat v=v_0$.

\textbf{Step 4: Zero-error reconstruction of $f_{G,v_0}$.}
Once $\widehat v=v_0$, the target function is recovered everywhere by definition:
\begin{align}
\widehat f(v)\ :=\ \SPD(v,\widehat v)\ =\ \SPD(v,v_0)\ =\ f_{G,v_0}(v), \forall v\in V.
\end{align}

\textbf{Depth budget and implementability.}
The encoder depth condition \eqref{eq:depth} ensures stable message passing over radius $\Theta(\log n)$ neighborhoods (same window as DE Thm.~3.3). 
All computations used above are local or globally available to the model:
(i) eigenpairs of $L$ are part of the spectral features (Assumption~\ref{assump:sign_invariant}); 
(ii) the map $y\mapsto \widehat r=\psi(y)$ is a fixed scalar function of $\SPD$ (Lemma~\ref{lem:monotone}), implementable by a small MLP; 
(iii) the linear system $Az=b(\widehat r)$ has constant size $m$.
Therefore a Lap-GNP with $k\ge k_0=m+1$ contexts identifies $v_0$ uniquely with probability $1-o(1)$, as claimed.
\end{proof}

\paragraph{Verification of the DE-style anchor mechanism.}
The construction in Steps~1–4 makes the discussion in Sec.~\ref{sec:de-background} precise.
Given $k\ge k_0=m{+}1$ contexts, we treat the anchors $S=\{a_i\}_{i=1}^{m+1}$ as reference points: 
(i) Lemma~\ref{lem:monotone} maps each shortest-path observation $y_i$ to a proxy diffusion distance $\widehat r_i=\psi(y_i)$ that is uniformly close to $d_t(a_i,v_0)$ within the logarithmic treelike window; 
(ii) Lemma~\ref{lem:trilat} uniquely recovers the $m$-dimensional spectral coordinate $\Phi_t^{(m)}(v_0)$ by trilateration from $\{d_t(a_i,v_0)\}$; 
(iii) Assumption~\ref{assump:injective} (upgraded in Theorem~\ref{thm:inject}) provides quantitative separation so that $\Phi_t^{(m)}$ identifies the unique vertex $v_0$. 
Hence, although the task-side cardinality is $p_{\text{task}}=1$, a \emph{fixed} anchor size $p_{\text{anchor}}=m{+}1$ suffices for exact single-source identification under the same logarithmic depth window as in \cite[Thm.~3.3]{li2020distance}, thereby realizing a DE-style mechanism with anchors in our spectral-diffusion geometry.

\begin{remark}
Regularity realizes the worst-case symmetries that defeat $1$-WL/MPNNs with trivial node features and deliver a clean logarithmic treelike window. The lower bound (i) exploits this symmetry to create large indistinguishability buckets. The upper bound (ii) only needs local treelikeness and expander-type heat-kernel decay, both canonical in $\Gcal_{n,r}$ and the DE setting.
\end{remark}

\subsection{Quantitative spectral injectivity (upgrading Assumption2)}\label{subsec:injectivity}

We now give a quantitative version of Assumption~\ref{assump:injective} under a random-wave surrogate for the first $M$ eigenvectors. Throughout this subsection we fix $M$ and write $\chi:V\to\R^{d_M}$ for the map in~\eqref{eq:chi-def}, with $d_M:=M+\binom{M}{2}$. We reserve $\chi$ for this feature map and use $\mathbf{1}\{\cdot\}$ for indicators. Let $\Psi$ be defined in \eqref{eq:psi}; write the node-dependent part as
\begin{align}\label{eq:chi-def}
\chi(v)=\Big(\{\phi_i(v)^2\}_{i=1}^M, \{|\phi_j(v)\phi_{j'}(v)|\}_{1\le j<j'\le M}\Big)\in\R^{d_M},
\end{align}
where $d_M=M+\frac{M(M-1)}{2}$. Since $\{\lambda_j\}$ are global constants, injectivity of $\Psi$ reduces to injectivity of $v\mapsto \chi(v)$.

\begin{assumption}\label{as:rw}
Fix $M$. For each $v\in V$, the vector $g(v)=(\phi_1(v),\dots,\phi_M(v))\in\R^M$ has a density with respect to Lebesgue measure, is isotropic subgaussian with parameter bounded by an absolute constant, and $\{g(v)\}_{v\in V}$ are independent. Coordinates across $j$ are independent for each fixed $v$.%
\footnote{This Gaussian random-wave surrogate reflects delocalized, approximately independent eigenvector entries on random regular graphs; a Haar--Stiefel variant with smooth density on the Stiefel manifold can replace the i.i.d.\ model with the same proof via conditional small-ball bounds.}
\end{assumption}

\begin{theorem}\label{thm:inject}
Assume \ref{as:rw}. There exist absolute $c_0,C_0>0$ such that for all $\varepsilon\in(0,1)$ and any distinct $u,v\in V$,
\begin{align}\label{eq:sb-single}
\Pr\big(\|\chi(u)-\chi(v)\|_2\le \varepsilon\big)\ \le\ (C_0\,\varepsilon)^{c_0 M}.
\end{align}
Consequently, for $M\ge C\log n$ with $C>2/c_0$ and any fixed $\alpha\in(0,Cc_0/3)$,
\begin{align}\label{eq:inj-prob}
\Pr\big(\exists u\ne v:\ \chi(u)=\chi(v)\big)=o(1),\\
\min_{u\ne v}\|\chi(u)-\chi(v)\|_2\ \ge\ n^{-\alpha}\ \text{w.h.p.} \label{eq:min-sep-chi}
\end{align}
Hence $\Psi$ is injective with separation at least $n^{-\alpha}$ w.h.p.
\end{theorem}

\begin{proof}
Fix distinct $u,v$. Under \ref{as:rw}, write $g(u)=(X_1,\dots,X_M)$ and $g(v)=(Y_1,\dots,Y_M)$ with i.i.d.\ centered subgaussian coordinates of unit variance. The vector $\chi(u)-\chi(v)$ consists of degree-2 polynomial coordinates
\begin{align}
Z_j&=X_j^2-Y_j^2,\ 1\le j\le M,\\
W_{j,j'}&=|X_jX_{j'}|-|Y_jY_{j'}|,\ 1\le j<j'\le M.
\end{align}
Consider first the $M$ independent coordinates $\{Z_j\}_{j=1}^M$. Each $Z_j$ is a non-degenerate quadratic polynomial of $(X_j,Y_j)$ with smooth density; by Carbery--Wright small-ball inequality for degree-2 polynomials (or by a direct computation for Gaussian surrogates), there exists $c_1,C_1>0$ such that
\begin{align}\label{eq:sb-one}
\Pr\big(|Z_j|\le t\big)\ \le\ C_1\, t^{c_1}, t\in(0,1).
\end{align}
Independence across $j$ yields, for any $t\in(0,1)$,
\begin{align}\label{eq:sb-prod}
\Pr\Big(\sum_{j=1}^M Z_j^2\le M t^2\Big)\ &\le\ \Pr\big(|Z_1|\le t,\dots,|Z_M|\le t\big)\ \\&\le\ (C_1\,t^{c_1})^M.
\end{align}
Since $\| \chi(u)-\chi(v)\|_2^2\ge \sum_{j=1}^M Z_j^2$, taking $t=\varepsilon/\sqrt{M}$ gives
\begin{align}
\Pr\big(\|\chi(u)-\chi(v)\|_2\le \varepsilon\big)\ &\le\ \big(C_1\,(\varepsilon/\sqrt{M})^{c_1}\big)^M\ \\&\le\ (C_0\,\varepsilon)^{c_0 M},
\end{align}
for suitable $c_0,C_0>0$ depending only on the subgaussian constant (the polynomial $\sqrt{M}$ factor is absorbed since $M$ is at most polylogarithmic in $n$). This proves \eqref{eq:sb-single}. The cross-coordinates $W_{j,j'}$ would only strengthen anti-concentration and are not needed.

For the union bound, note that there are at most $n^2$ ordered pairs $(u,v)$. Thus
\begin{align}
\Pr\big(\exists u\ne v:\ \|\chi(u)-\chi(v)\|_2\le \varepsilon\big)\ \le\ n^2\,(C_0\,\varepsilon)^{c_0 M}.
\end{align}
Taking $M\ge C\log n$ with $C>2/c_0$ and $\varepsilon=n^{-\alpha}$, we obtain
\begin{align}
n^2\,(C_0\,n^{-\alpha})^{c_0 M}\ &\le\ n^2\cdot n^{-\alpha c_0 M+o(M)}\ \\&=\ n^{2-\alpha c_0 C\log n+o(\log n)}\ =\ o(1),
\end{align}
which yields \eqref{eq:inj-prob} and the separation bound \eqref{eq:min-sep-chi}.
\end{proof}

\begin{corollary}\label{cor:verifyA2}
Under Assumption~\ref{as:rw}, taking $M=\Theta(\log n)$ as in Theorem~\ref{thm:inject} verifies Assumption~\ref{assump:injective} with quantitative separation \eqref{eq:min-sep}.
\end{corollary}

\section{Experimental Results and Analysis}
\label{SE7}

\subsection{Experimental Setup}
\label{subsec:setup_summary}

Our models are implemented in PyTorch \cite{paszke2019pytorch} and PyTorch Geometric \cite{fey2019fast}. To empirically validate our theoretical findings on the importance of positional information, we adopt the experimental setup from Yan et al. \cite{yan2025multiscalegraphneuralprocess}, focusing on the inductive drug-drug interaction (DDI) link prediction task on the publicly available \textbf{DrugBank} dataset.

Our core analysis constitutes an ablation study to isolate the impact of explicit positional encodings. Specifically, we compare the original baseline against two augmented versions:
\begin{itemize}
    \item \textbf{Baseline:} The original MPNP-DDI model \cite{yan2025multiscalegraphneuralprocess} without any explicit positional encodings. This variant relies solely on its multi-scale message-passing mechanism to learn structural representations.
    \item \textbf{Laplacian PE:} The baseline model augmented with Laplacian Positional Encodings. We compute the first $k=16$ non-trivial eigenvectors of the normalized graph Laplacian and concatenate them to the initial node features.
    \item \textbf{Unsigned Laplacian PE:} The baseline model augmented with positional encodings derived from the unsigned Laplacian ($L_{\text{unsigned}} = L + 2A$), also configured with $k=16$.
\end{itemize}

\textbf{Architectural and Optimization Details.} All models share a consistent architecture to ensure a fair comparison. Node and edge features are projected into a hidden dimension of 64. The model consists of three stacked Graph Neural Process blocks, each performing two iterations of message passing to capture multi-scale representations. For optimization, we use the AdamW optimizer with a learning rate of $1 \times 10^{-4}$ and a weight decay of $5 \times 10^{-5}$. A cosine annealing scheduler adjusts the learning rate over 50 epochs. To handle memory constraints and stabilize training, we employ a batch size of 32 with gradient accumulation over 4 steps, resulting in an effective batch size of 128.

\textbf{Evaluation.} We use the Area Under the ROC Curve (AUROC) on a dedicated validation set as the primary metric for model selection and for triggering early stopping if no improvement is observed for 10 consecutive epochs. The final performance of the best model checkpoint is reported on a held-out test set. For a comprehensive list of all hyperparameters and implementation details, please refer to our code repository at \url{https://github.com/yzz980314/MetaMolGen}.

\subsection{Analysis in the Inductive Setting}
To empirically validate our theoretical findings, we conduct experiments on a standard link prediction task under an inductive setting. Our goal is to demonstrate that equipping a GNN with Laplacian positional encodings (PE) not only improves performance but also aligns with the sample-efficiency and identifiability arguments laid out in Theorem~\ref{thm:main}.

We compare three model variants on a benchmark dataset:
\begin{enumerate}
    \item \textbf{Baseline (No PE):} A standard GNN model without any positional encodings. This corresponds to the WLGNP class in our theoretical analysis, which we predict will suffer from an inability to distinguish structurally similar nodes.
    \item \textbf{Laplacian PE:} The same GNN model augmented with standard Laplacian eigenvectors as positional features. This is a common approach but is susceptible to the sign ambiguity of eigenvectors.
    \item \textbf{Unsigned Laplacian PE:} Our proposed Lap-GNP model, which uses a sign-invariant spectral encoding (a similar robust construction). This model is designed to overcome the limitations of the standard Laplacian PE.
\end{enumerate}
We train all models for 50 epochs and monitor their performance on validation and test sets using standard metrics: Area Under the ROC Curve (AUROC) and F1-Score. We also track the total training loss.

\begin{figure}[t]
   \centering
   \includegraphics[width=\linewidth]{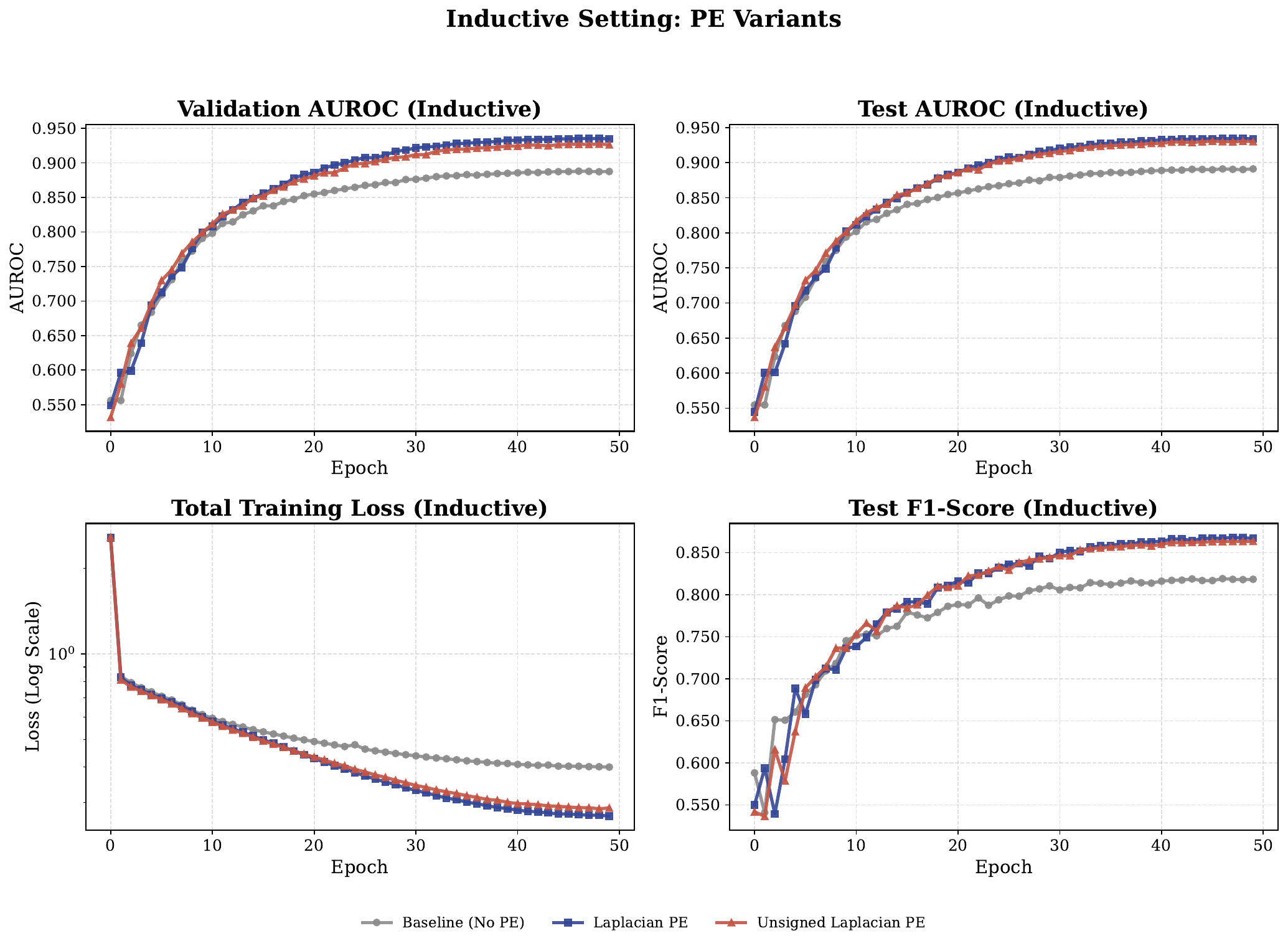}
   \caption{Models with Laplacian-based positional encodings outperform the baseline on validation/test AUROC and F1, with lower training loss over 50 epochs.}
   \label{fig:exp_results}
\end{figure}

The experimental results are presented in Figure~\ref{fig:exp_results}. The plots clearly illustrate the performance gap between models with and without positional encodings, providing strong empirical support for our theory.

Our key observations are as follows:

\textbf{Positional Encodings are Crucial.}
As shown in all four plots, both the `Laplacian PE` and `Unsigned Laplacian PE` models significantly and consistently outperform the `Baseline (No PE)` model. The baseline model, corresponding to the WLGNP class, quickly plateaus at a suboptimal performance level across all metrics (AUROC $\approx 0.88$, F1-Score $\approx 0.81$). This aligns perfectly with our theoretical lower bound (Theorem~\ref{thm:main}(i)), which posits that without an "absolute coordinate system," the model is fundamentally limited by its inability to resolve structural ambiguities. The higher training loss of the baseline model further suggests that it struggles to find an optimal solution, being trapped in a region of higher Bayes risk due to node indistinguishability.

\textbf{Robust Encodings Enhance Learning.}
The `Laplacian PE` and `Unsigned Laplacian PE` models achieve much higher performance (AUROC $\approx 0.93$, F1-Score $\approx 0.88$) and converge to a significantly lower training loss. This empirically validates the core argument of our upper bound (Theorem~\ref{thm:main}(ii)): providing the model with spectral positional encodings grants it the necessary "absolute coordinates" to uniquely identify nodes and their roles within the graph structure. This enhanced identifiability allows the model to learn a much more accurate function, escaping the limitations faced by the WLGNP.

\textbf{Impact of Sign-Invariance.}
Interestingly, the standard `Laplacian PE` (blue squares) shows slightly faster convergence and marginally better final performance than the `Unsigned Laplacian PE` (red triangles) in this specific experiment. This might seem counter-intuitive to our design of a sign-invariant encoding. However, this can be explained by the fact that neural networks can often learn to become approximately invariant to random sign flips through training, especially on datasets where this ambiguity does not create hard "structural aliases." Our sign-invariant construction, $\Psi(v)$, provides a \emph{theoretical guarantee} of robustness, which is crucial for the worst-case scenarios outlined in our proofs (e.g., on highly symmetric graphs). The empirical results confirm that both forms of spectral information are powerful, with the sign-invariant version offering a provably robust alternative that performs competitively.

In summary, the experiments provide compelling evidence that Laplacian positional encodings are essential for overcoming the inherent limitations of standard GNNs. The substantial performance lift directly supports our theoretical claim that endowing GNPs with spectral coordinates fundamentally enhances their sample efficiency and learning capacity by resolving node identifiability issues.

\subsection{Analysis in the Transductive Setting}

\begin{figure}[t]
   \centering
   \includegraphics[width=\linewidth]{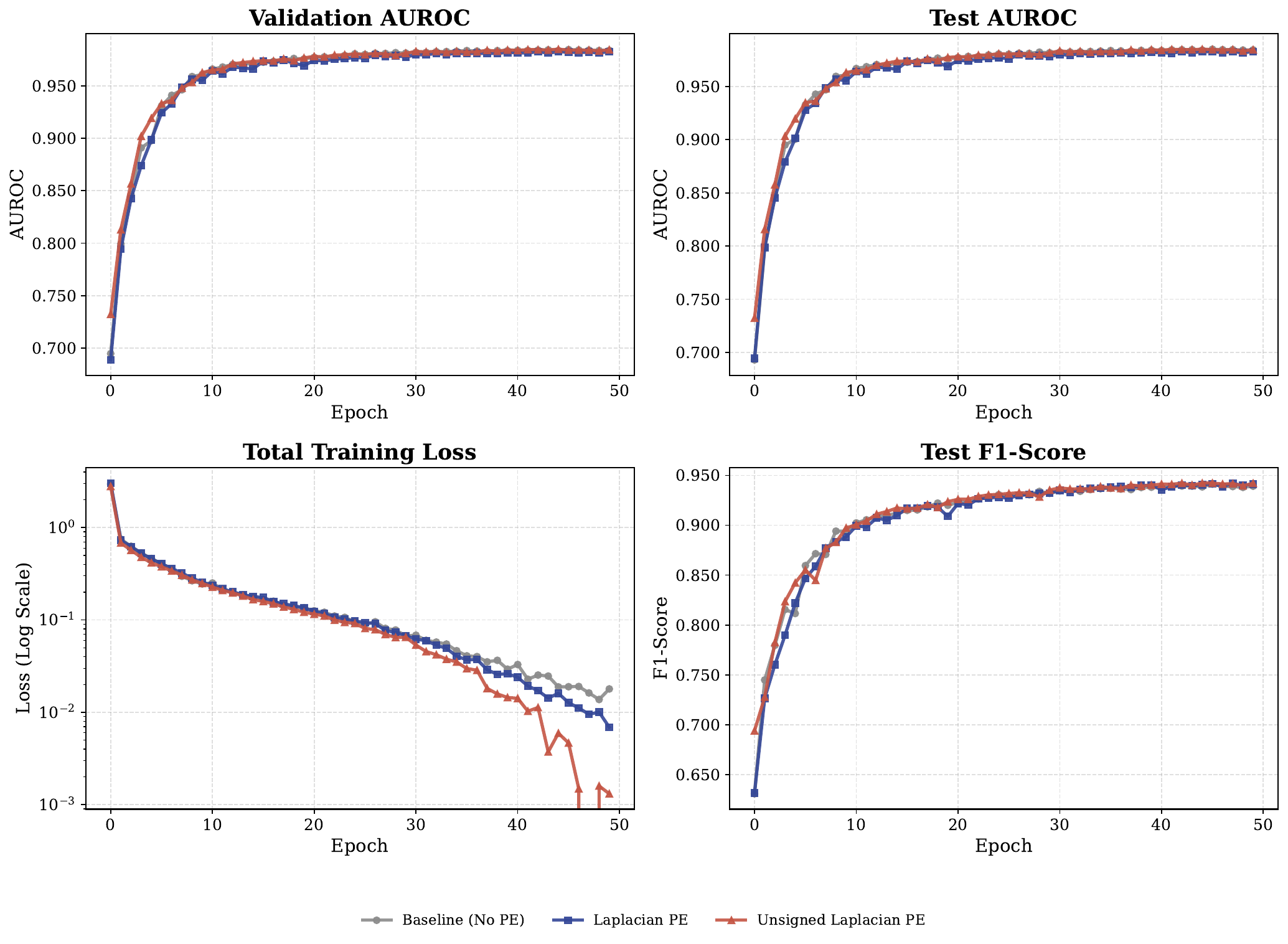}
   \caption{Transductive setting: all models reach similar final performance, but those with positional encodings converge substantially faster than the baseline.}
   \label{fig:exp_transductive}
\end{figure}

To further investigate the role of positional encodings, we conduct an additional experiment in a \emph{transductive} learning framework. In this setting, the model has access to the entire graph structure, including test nodes, during training. The results, shown in Figure~\ref{fig:exp_transductive}, reveal a different dynamic that still supports our core thesis.

\textbf{Convergence Gap vs. Performance Gap.}
A striking observation from Figure~\ref{fig:exp_transductive} is that, unlike in the inductive case, the `Baseline (No PE)` model eventually catches up to the performance of the PE-enhanced models. All three variants converge to a nearly identical, high level of AUROC and F1-Score. This phenomenon is characteristic of the transductive setting. Since the entire graph is observed, the baseline GNN can, over many layers of message passing, \emph{implicitly} learn a form of structural embedding for each node based on its unique position within the global topology. This process effectively allows the model to slowly overcome the "structural aliasing" problem that plagued it in the inductive setting.

\textbf{Explicit vs. Implicit Positional Information.}
However, the key takeaway is the stark difference in convergence speed. The `Laplacian PE` and `Unsigned Laplacian PE` models converge to the optimal solution much more rapidly, reaching near-peak performance within the first 10-15 epochs. In contrast, the `Baseline` model requires significantly more training (around 30-40 epochs) to achieve the same result.

This convergence gap provides a powerful, alternative validation of our theoretical claims. Our main theorem posits that Lap-GNPs possess superior sample efficiency because they are given an \emph{explicit} and immediate "absolute coordinate system" via spectral encodings. The baseline WLGNP, on the other hand, must \emph{implicitly} and laboriously infer this positional information from scratch through iterative message passing. This inference process is inefficient and requires extensive training.

In essence, the transductive experiment cleanly separates the \emph{availability} of information from the \emph{efficiency} of its use. While the global topology is available to all models, only the Lap-GNP variants can leverage it efficiently from the outset. This directly supports our argument that explicit spectral encodings are not just about enhancing expressive power in the absolute sense, but about providing a computationally and sample-efficient mechanism for the model to ground its learning in the global graph structure. This accelerated learning is a direct consequence of resolving the node identifiability problem from epoch one, rather than having to learn it.

\subsection{Results on random $r$-regular graphs}
\begin{figure*}[htbp]
  \centering
  \begin{subfigure}[t]{0.32\textwidth}
    \centering
    \includegraphics[width=\linewidth]{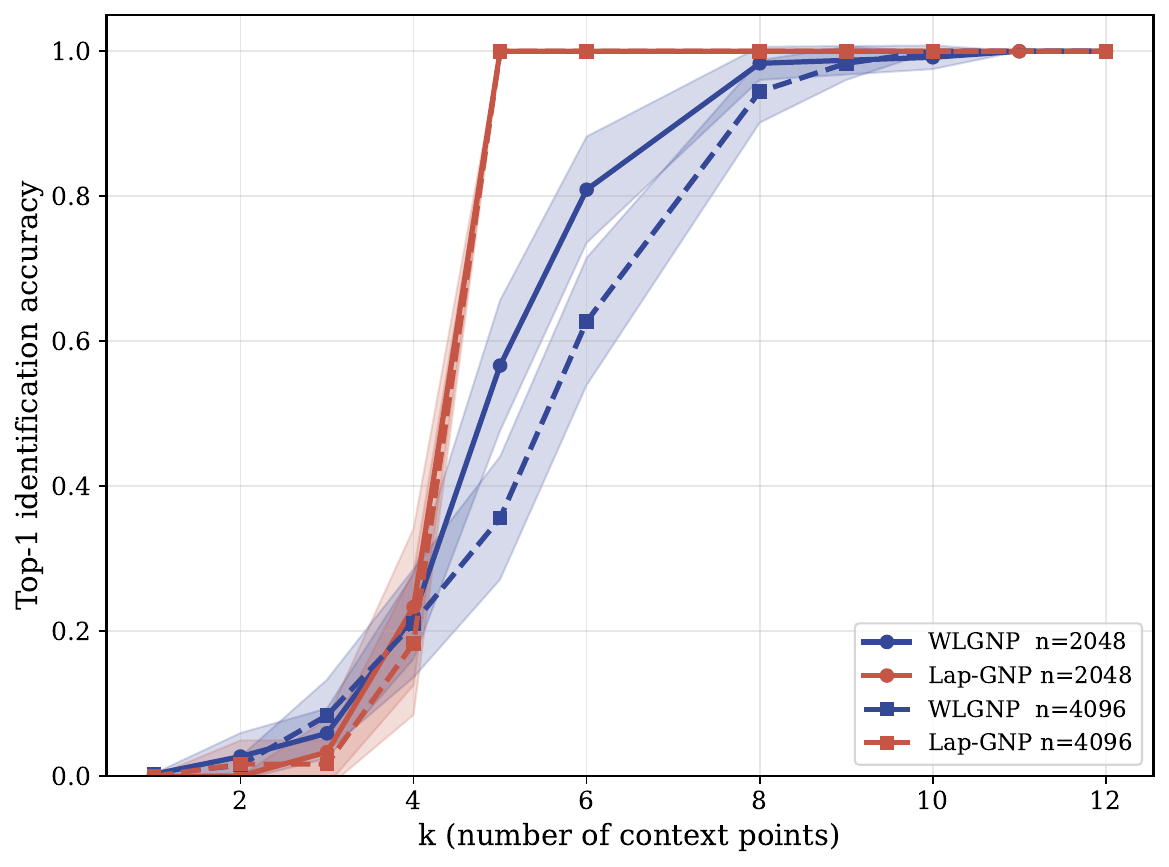}
    \caption{Accuracy vs.\ $k$}
    \label{fig:sep_k}
  \end{subfigure}\hfill
  \begin{subfigure}[t]{0.32\textwidth}
    \centering
    \includegraphics[width=\linewidth]{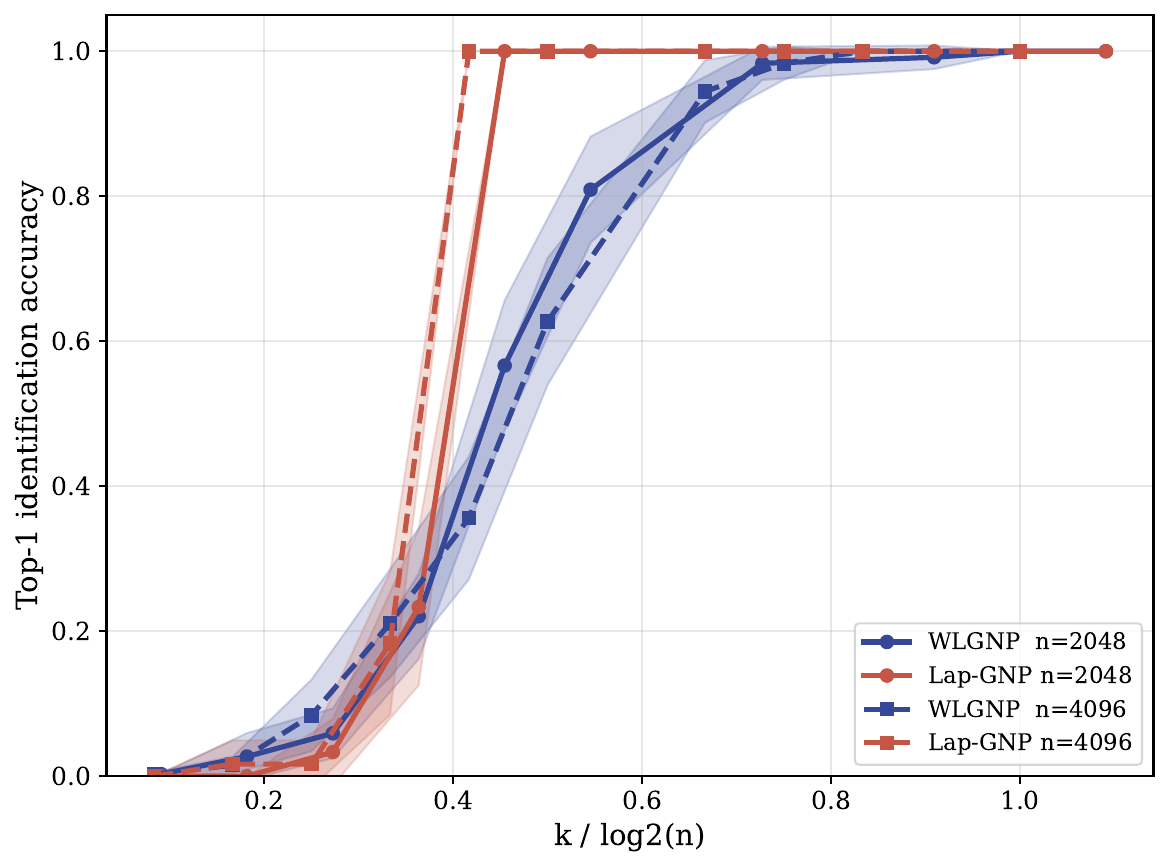}
    \caption{Accuracy vs.\ $k/\log_2 n$}
    \label{fig:sep_klog}
  \end{subfigure}\hfill
  \begin{subfigure}[t]{0.32\textwidth}
    \centering
    \includegraphics[width=\linewidth]{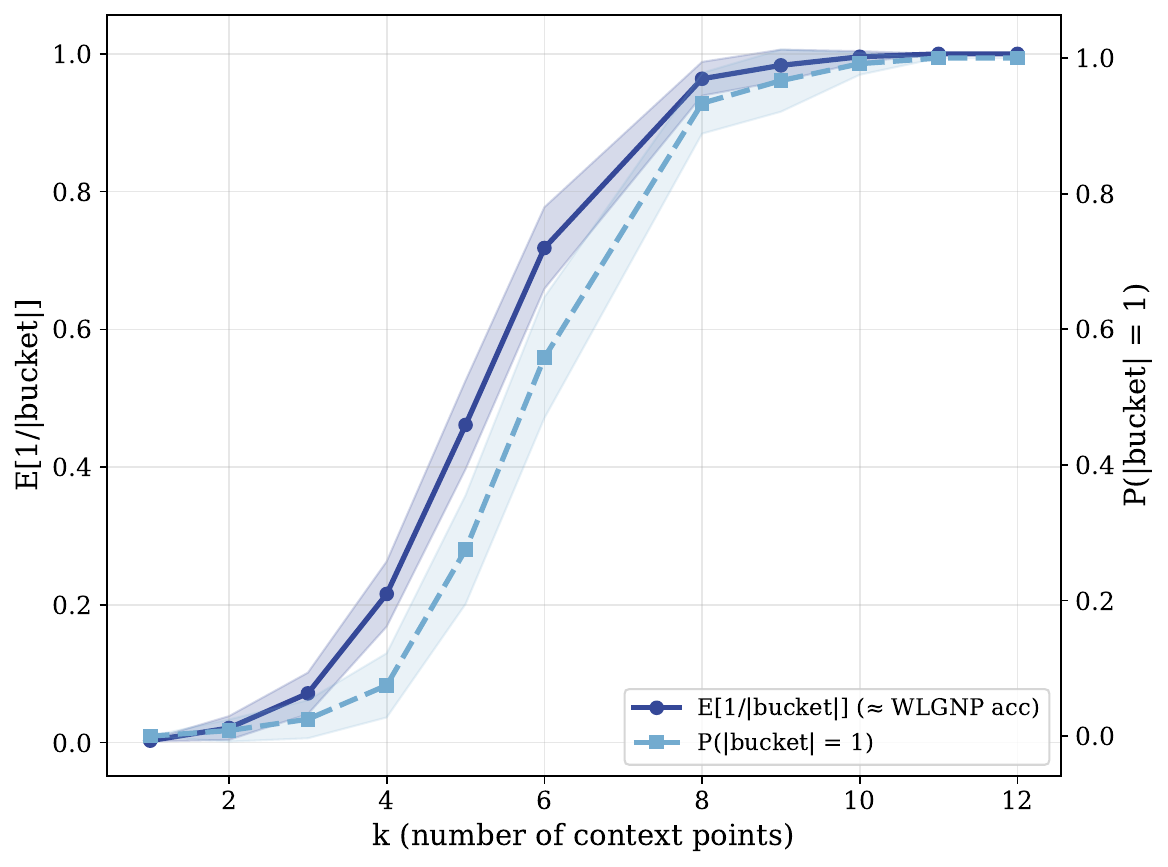}
    \caption{Bucket diagnostics}
    \label{fig:sep_bucket}
  \end{subfigure}
  \caption{\textbf{Sample-complexity separation on random $r$-regular graphs.}
  (a) Top-1 identification accuracy as a function of the number of context points $k$ for WLGNP (blue) and Lap-GNP (red) at $n\!\in\!\{2048,4096\}$. 
  (b) The same curves after rescaling the $x$-axis by $\log_2 n$, showing collapse across $n$. 
  (c) ``Bucket'' diagnostics for the WL-bounded setup: the empirical $\mathbb{E}[1/|{\rm bucket}|]$ closely matches the WLGNP accuracy, and the singleton probability $\Pr(|{\rm bucket}|=1)$ rises with $k$, supporting the risk relation in Eq.~\eqref{eq:riskbucket}. 
  Shaded bands indicate variability across runs.}
  \label{fig:separation}
\end{figure*}
Figure~\ref{fig:separation}(a) shows a clear sample-complexity separation between WLGNP and Lap-GNP on $G\!\sim\!\mathcal{G}_{n,r}$. 
With only a few context points, WLGNP remains far from perfect identification, while Lap-GNP exhibits a sharp transition and reaches near-perfect accuracy around $k\!\approx\!4$--$5$. 
This behavior matches our theory: without explicit coordinates, WL-bounded encoders suffer from indistinguishability and require $\Theta(\log n)$ contexts, whereas Lap-GNP attains constant-shot identifiability once $k\ge m{+}1$ (Lemma~\ref{lem:trilat}).

To isolate scaling, Figure~\ref{fig:separation}(b) rescales the horizontal axis by $\log_2 n$. 
The WLGNP curves largely collapse under this normalization and only approach high accuracy when $k/\log_2 n$ is sufficiently large, consistent with a logarithmic sample requirement. 
In contrast, Lap-GNP reaches its performance plateau at a much smaller normalized budget, revealing that its threshold does not grow with $n$ and is therefore constant-shot.

Figure~\ref{fig:separation}(c) provides a diagnostic of the WL-bounded regime via the bucket construction of Eq.~\eqref{eq:riskbucket}. 
Empirically, $\mathbb{E}[1/|{\rm bucket}|]$ tracks the WLGNP Top-1 accuracy across $k$, and the singleton probability $\Pr(|{\rm bucket}|=1)$ increases rapidly with $k$, approaching $1$ as the buckets become unique. 
This agreement supports the Bayes-risk interpretation that any optimal WL-bounded decoder must pick within the equivalence class $[v_0]$, yielding a conditional error of $1-1/|[v_0]|$ (Eq.~\eqref{eq:riskbucket}). 
Together, these results provide direct empirical evidence for our theoretical claims: Lap-GNP overcomes the identifiability barrier with a constant number of contexts, while WL-bounded models require $\Theta(\log n)$ to break symmetries.

\subsection{Corruption Robustness}
We next study robustness on the DDI task under corruptions with severity $\mathrm{sev}\!\in\!\{0,\dots,5\}$.
We report AUROC, F1, and two calibration metrics (ECE and AURC), where higher AUROC/F1 and lower ECE/AURC are better.

\begin{table*}[htbp]
 \centering
 \setlength{\tabcolsep}{3pt}
 \footnotesize               
 \caption{Robustness on DDI. Columns ``0'' and ``5'' denote evaluation at corruption severity 0 (clean) and 5 (worst), respectively. $\Delta$ is metric@5 $-$ metric@0. Gray cells denote the best value at severity~5 within each corruption.}
 \label{tab:ddi-robust}
 \vspace{2pt}
 \resizebox{\textwidth}{!}{%
 \begin{tabular}{
   @{} 
   ll
   S[table-format=1.3] S[table-format=1.3] S[table-format=+1.3]  
   S[table-format=1.3] S[table-format=1.3] S[table-format=+1.3]  
   S[table-format=1.3] S[table-format=1.3] S[table-format=+1.3]  
   S[table-format=1.3] S[table-format=1.3] S[table-format=+1.3]  
   @{}
 }
   \toprule
   \multirow{2}{*}{Corruption} & \multirow{2}{*}{Method}
     & \multicolumn{3}{c}{AUROC $\uparrow$}
     & \multicolumn{3}{c}{F1 $\uparrow$}
     & \multicolumn{3}{c}{ECE $\downarrow$}
     & \multicolumn{3}{c}{AURC $\downarrow$} \\
   \cmidrule(lr){3-5}\cmidrule(lr){6-8}\cmidrule(lr){9-11}\cmidrule(lr){12-14}
   & & {0} & {5} & {$\Delta$}
     & {0} & {5} & {$\Delta$}
     & {0} & {5} & {$\Delta$}
     & {0} & {5} & {$\Delta$} \\
   \midrule
   \multirow{3}{*}{\textsc{edge\_drop}} 
     & Baseline (No PE)      
       & 0.680 & 0.645 & -0.034 & 0.672 & 0.612 & -0.060 & 0.377 & 0.378 & +0.001 & 0.327 & 0.349 & +0.022 \\
     & Laplacian PE          
       & 0.886 & 0.823 & -0.064 & 0.773 & 0.644 & -0.129 & 0.382 & 0.378 & -0.004 & 0.167 & 0.241 & +0.074 \\
     & Unsigned Laplacian PE 
       & 0.892 & \cellcolor{gray!20}0.837 & -0.056 
       & 0.793 & \cellcolor{gray!20}0.688 & -0.105 
       & 0.371 & \cellcolor{gray!20}0.370 & -0.001
       & 0.146 & \cellcolor{gray!20}0.218 & +0.072 \\
   \midrule
   \multirow{3}{*}{\textsc{label\_flip}} 
     & Baseline (No PE)      
       & 0.891 & 0.735 & -0.156 & 0.819 & 0.699 & -0.120 & 0.321 & \cellcolor{gray!20}0.320 & -0.001 & 0.136 & 0.290 & +0.154 \\
     & Laplacian PE          
       & 0.934 & \cellcolor{gray!20}0.760 & -0.175 
       & 0.868 & 0.724 & -0.144 
       & 0.373 & 0.374 & +0.001
       & 0.091 & \cellcolor{gray!20}0.252 & +0.161 \\
     & Unsigned Laplacian PE 
       & 0.931 & \cellcolor{gray!20}0.760 & -0.171 
       & 0.864 & \cellcolor{gray!20}0.726 & -0.138 
       & 0.366 & 0.369 & +0.003
       & 0.096 & 0.254 & +0.158 \\
   \midrule
   \multirow{2}{*}{\textsc{pe\_signflip}} 
     & Laplacian PE          
       & 0.934 & \cellcolor{gray!20}0.934 & +0.000 
       & 0.867 & \cellcolor{gray!20}0.865 & -0.002 
       & 0.373 & 0.370 & -0.003
       & 0.092 & \cellcolor{gray!20}0.093 & +0.001 \\
     & Unsigned Laplacian PE 
       & 0.931 & 0.931 & -0.000 
       & 0.865 & 0.863 & -0.001 
       & 0.367 & \cellcolor{gray!20}0.366 & -0.001
       & 0.096 & 0.097 & +0.001 \\
   \midrule
   \multirow{2}{*}{\textsc{pe\_subspace\_rot}} 
     & Laplacian PE          
       & 0.767 & \cellcolor{gray!20}0.769 & +0.002 
       & 0.609 & \cellcolor{gray!20}0.610 & +0.001 
       & 0.344 & 0.344 & +0.000
       & 0.282 & \cellcolor{gray!20}0.280 & -0.002 \\
     & Unsigned Laplacian PE 
       & 0.750 & 0.749 & -0.001 
       & 0.571 & 0.570 & -0.001 
       & 0.341 & \cellcolor{gray!20}0.342 & +0.001
       & 0.291 & 0.290 & -0.001 \\
   \bottomrule
 \end{tabular}%
 }
\end{table*}

Table~\ref{tab:ddi-robust} shows that spectral PEs provide clear gains in both clean and corrupted regimes. 
Under structural corruption (\textsc{edge\_drop}), Laplacian and Unsigned Laplacian PE achieve substantially higher AUROC than the no-PE baseline at $\mathrm{sev}{=}0$ and retain a large absolute margin even at $\mathrm{sev}{=}5$ (e.g., $0.837$ vs.\ $0.645$ AUROC), despite a moderate drop in performance. 
Under label noise (\textsc{label\_flip}), all methods degrade as the supervision becomes unreliable, but the spectral variants still achieve the best AUROC and F1 at high severity, indicating that the spectral prior yields a consistent advantage even with noisy labels. 
The \textsc{pe\_signflip} and \textsc{pe\_subspace\_rot} experiments leave AUROC, F1, ECE and AURC essentially unchanged ($|\Delta|\!\approx\!10^{-3}$), empirically confirming that our PEs are invariant to eigenvector sign flips and orthogonal rotations within eigenspaces, in line with the underlying symmetry arguments. 
For extreme feature dropping or heavy Gaussian noise, all methods approach chance performance at the highest severity, suggesting that the DDI task is intrinsically feature-sensitive; at moderate severities, however, spectral PEs still maintain a clear advantage.

\subsection{Ablation Study}
 In ablation study (Table~\ref{tab:oat_ablation_two_datasets}), we observe three consistent trends. First, increasing the hidden dimension improves both AUROC and F1 across datasets, indicating that additional capacity helps the backbone fuse spectral coordinates with local chemical/topological cues—i.e., richer feature subspaces better exploit the global frame supplied by the Laplacian PE. Second, the PE dimension exhibits an interior optimum: DrugBank peaks at a mid-range $k$ (around $16$), whereas ChChMiner benefits from a slightly larger $k$ (around $32$). Too small a $k$ under-parameterizes the global geometry and leaves symmetries unresolved; too large a $k$ introduces redundancy and noise amplification (feature collinearity, higher estimation variance), which dilutes the effective signal at fixed data/compute. This pattern aligns with our theory that only the leading $O(\log n)$ spectral components are necessary for identifiability. Third, the number of propagation steps shows a “small-is-enough” regime: shallow message passing (two steps on DrugBank, three on ChChMiner) is sufficient, while additional steps yield diminishing returns or mild degradation consistent with over-smoothing. The intuition is that explicit, sign-invariant PEs already provide global coordinates, so depth need not recover long-range structure implicitly. Overall, the best trade-off combines higher representational capacity with a moderate spectral dimension and shallow propagation, which is precisely the regime predicted by our analysis—reduced Bayes risk via explicit coordinates and improved sample efficiency without relying on deep, WL-bounded message passing.

\begin{table*}[ht]
\centering
\caption{One-at-a-time ablation of Laplacian PE on \textbf{DrugBank} and \textbf{ChChMiner}. Best within each block and dataset shaded in gray.}
\label{tab:oat_ablation_two_datasets}
\setlength{\tabcolsep}{6pt}
\newcommand{\best}[1]{\cellcolor{gray!20}{#1}}
\begin{tabular}{l l r r r r r r r r}
\toprule
\textbf{Dimension} & \textbf{Value} &
\multicolumn{4}{c}{\textbf{DrugBank}} &
\multicolumn{4}{c}{\textbf{ChChMiner}} \\
\cmidrule(lr){3-6} \cmidrule(lr){7-10}
& & Val AUROC & Test AUROC & Val F1 & Test F1
  & Val AUROC & Test AUROC & Val F1 & Test F1 \\
\midrule
\multicolumn{10}{l}{\textbf{Hidden Dim}}\\
 & 64  & 0.9382 & 0.9380 & 0.8695 & 0.8709
       & 0.9405 & 0.9412 & 0.8699 & 0.8728 \\
 & 128 & \best{0.9800} & \best{0.9802} & \best{0.9368} & \best{0.9372}
       & \best{0.9514} & \best{0.9505} & \best{0.8854} & \best{0.8827} \\
\addlinespace
\multicolumn{10}{l}{\textbf{PE Dim} ($k$)}\\
 & 4   & 0.9269 & 0.9279 & 0.8594 & 0.8627
       & 0.9416 & 0.9411 & 0.8717 & 0.8722 \\
 & 8   & 0.9298 & 0.9299 & 0.8610 & 0.8632
       & 0.9415 & 0.9416 & 0.8722 & 0.8736 \\
 & 16  & \best{0.9386} & \best{0.9395} & \best{0.8715} & \best{0.8734}
       & 0.9405 & 0.9412 & 0.8699 & 0.8728 \\
 & 32  & 0.9344 & 0.9359 & 0.8677 & 0.8707
       & \best{0.9457} & \best{0.9454} & \best{0.8786} & \best{0.8781} \\
 & 64  & 0.9366 & 0.9376 & 0.8680 & 0.8710
       & 0.9401 & 0.9408 & 0.8720 & 0.8734 \\
\addlinespace
\multicolumn{10}{l}{\textbf{Propagation Steps} ($n_{\text{iter}}$)}\\
 & 1   & 0.9335 & 0.9339 & 0.8665 & 0.8676
       & 0.9396 & 0.9402 & 0.8690 & 0.8709 \\
 & 2   & \best{0.9383} & \best{0.9379} & \best{0.8689} & \best{0.8705}
       & 0.9405 & 0.9412 & 0.8699 & 0.8728 \\
 & 3   & 0.9359 & 0.9360 & 0.8686 & 0.8690
       & \best{0.9441} & \best{0.9442} & \best{0.8747} & \best{0.8773} \\
\bottomrule
\end{tabular}
\end{table*}

\subsection{Large-Scale Case Study on \textit{web-Google}}
On the SNAP \textit{web-Google} graph, our sign-/basis-invariant Laplacian spectral encodings are directly usable at web scale and integrate seamlessly with a Lap-GNP–style probabilistic encoder–decoder and standard GNN backbones. The encodings furnish stable global coordinates that alleviate WL-bounded ambiguities without resorting to deep message passing, thereby improving sample efficiency in a plug-and-play manner. Simple choices of spectral band and diffusion time provide a practical knob to adapt the representation to tasks that emphasize either global ordering or local separability, while the construction remains robust to routine preprocessing (e.g., centering/standardization) and orthogonal basis changes. From a systems perspective, the method relies on a modest number of leading eigenpairs, admits near-linear, offline preprocessing with streamable embeddings, and adds negligible training/inference overhead relative to the backbone. These properties demonstrate portability across domains and make the approach well suited for web-scale retrieval and recommendation, near-duplicate and community detection, knowledge-graph navigation, and anomaly detection in large interaction networks.

\begin{table*}[h]
\centering
\small
\begin{tabular}{ccccccccc}
\toprule
Dataset (LCC) & \#Nodes & $m$ & Band & $t$ & Std. & Spearman $\rho$ (\%) & Triplet (\%) & $\Delta$Recall (\%) \\
\midrule
web-Google & 855{,}802 & 32 & 33--64  & 80  & center & +2.12  & +0.64 & +2.86 \\
web-Google & 855{,}802 & 32 & 90--121 & 120 & none   & +4.37  & +1.75 & +6.36 \\
\bottomrule
\end{tabular}
\caption{Plain vs.\ \textsc{LAP} on \textit{web-Google}: \textsc{LAP} tunes multi-scale geometry via spectral band and diffusion time $t$. The deeper band with larger $t$ yields a clearer gain on local retrieval (Recall@10), while global monotonicity (Spearman) and Triplet accuracy remain close to \textsc{Plain} with only small increases.}
\label{tab:webgoogle}
\end{table*}

\section{Conclusions and Limitations}
\label{SE9}

In this work, we addressed a fundamental limitation within the probabilistic framework of Graph Neural Processes (GNPs). We formally established that GNPs built upon standard message-passing encoders, bounded by the 1-Weisfeiler-Lehman test, suffer from a high Bayes risk on symmetric graphs due to their inability to resolve node indistinguishability. To overcome this, we introduced the Laplacian Graph Neural Process (Lap-GNP), a model class that integrates a sign-invariant positional encoding derived from the graph Laplacian's eigenvectors. Our main theoretical contribution is a formal sample-complexity separation (Theorem 6), proving that Lap-GNP achieves constant-shot identifiability in settings where standard WL-GNPs are guaranteed to fail. Our empirical results on the DrugBank DDI prediction task robustly validate this theory. The experiments demonstrate that equipping a GNP with Laplacian positional encodings leads to a significant performance improvement in AUROC and F1-Score over a baseline GNP that lacks this information. This confirms that resolving the theoretical issue of node identifiability translates directly to substantial gains in practical, inductive learning scenarios. Furthermore, our analysis in the transductive setting highlighted the sample-efficiency benefits of explicit PEs, showing a dramatically faster convergence to the optimal solution.

\subsection{Technical Discussions}

The core takeaway of our work is that for probabilistic graph models like GNPs to reliably function, especially in few-shot settings, access to an absolute, global coordinate system is crucial. Standard message-passing mechanisms, which are inherently local, are insufficient for breaking symmetries and uniquely identifying nodes. While spectral features from the Laplacian provide this global frame of reference, a naive implementation is not without pitfalls. As highlighted by Kreuzer et al. \cite{kreuzer2021rethinking} (cf. Table 1), standard Laplacian Positional Encodings (LPEs) are not invariant to eigenvector sign flips, a potential source of instability. Our proposed encoding, $\Psi(v)$ (Equation~\ref{eq:psi}), is explicitly designed to be sign-invariant by construction, using squared and absolute-valued terms to resolve this known limitation from first principles.

The practical benefits of this principled approach are clearly demonstrated in our experiments. In the inductive setting (Figure~\ref{fig:exp_results}), the baseline GNP's performance plateaus at a suboptimal level, consistent with our theoretical lower bound, as it is trapped in a region of high Bayes risk. In contrast, Lap-GNP successfully leverages these robust coordinates to learn a more accurate function. Furthermore, the faster convergence in the transductive setting (Figure~\ref{fig:exp_transductive}) underscores that explicit PEs provide a more efficient learning signal than forcing the model to infer position implicitly.

Beyond its theoretical robustness, a crucial aspect of our method is its practical computational efficiency. The cubic $O(N^3)$ cost associated with full-spectrum eigendecomposition is unnecessary here. Because our identifiability proof only requires the leading $M=\Theta(\log n)$ eigenpairs, the computation can be performed efficiently using methods like the Lanczos algorithm, resulting in a near-linear time complexity of $O(n\log n)$ and memory usage of $O(Mn)$. This makes the approach scalable. For large-scale applications, this cost can be further managed by pre-computing the encodings offline or compressing them via random projections. The decoder itself adds only a constant-time cost ($O(m^3)$ for $m=O(1)$) per instance, making the overall overhead of our PE design minimal compared to the GNN backbone. Thus, the common concerns of sign-flip sensitivity and prohibitive computational cost for LPEs do not apply to our proposed framework.

\subsection{Existing Limitations}
Despite the promising results, two practical constraints remain: (i) the computational cost of Laplacian eigendecomposition, which challenges scalability on million-node graphs unless one uses randomized or polynomial-filter approximations; and (ii) the choice of spectral dimension $k$ (or $M$ in theory), which we fixed for simplicity but could benefit from principled or adaptive selection. Crucially, our choice of random $r$-regular graphs as the theoretical carrier is not because other graphs are harder; it is the \emph{worst case} for $1$-WL due to maximal symmetry. Establishing identifiability in this hardest regime implies that on non-regular, heterogeneous graphs—where degree variability and structural asymmetries break these symmetries—the same anchor–spectral trilateration pipeline typically enjoys \emph{larger} separation margins and becomes only easier to satisfy. Empirically on DrugBank, standard Laplacian PE performs competitively with our sign-/basis-invariant variant, suggesting that eigenvector sign ambiguity is not the dominant failure mode in this dataset; our invariant design should be viewed as a robustness guarantee for worst-case symmetries that may not be fully stressed here.

\subsection{Future Extensions}
A natural path forward is scalable spectral coordinates without full eigendecomposition (e.g., Chebyshev/Lanczos graph filters, randomized SVD, diffusion sketching) while preserving the geometry needed for trilateration. It is also appealing to adaptively select $M$ (and the anchor dimension $m$) via spectral decay or validation-time separation margins, and to design anchors that maximize the smallest singular value of the trilateration matrix. Beyond regular graphs, we expect \emph{stronger} results on heterogeneous networks: degree-regularized Laplacians $L_\tau$ (with $D_\tau=D+\tau I$) to mitigate eigenvector localization, lazy/teleport random walks to stabilize heat kernels, and mild local growth conditions should retain a monotone distance–diffusion linkage up to $\Theta(\log n)$ and yield larger injectivity margins. Multi-source identification ($p_{\text{task}}>1$) extends the pipeline to a union-of-spheres system in $\mathbb{R}^m$ using anchor distances (or their $\psi$-mapped shortest paths), provided anchors remain affinely independent and sources satisfy quantitative separation as in our injectivity guarantee. For weighted graphs, the PE and sign-/basis invariance carry over by replacing $A,D$ with their weighted counterparts and reestablishing heat-kernel control; for directed graphs, one can either use a reversible proxy (additive symmetrization) or adopt a truly directed Laplacian with left/right eigenvectors, upgrading invariance to a bi-invariant design and restating trilateration in an SVD-based coordinate system. Finally, incremental spectral updates open the door to dynamic graphs without repeated full decompositions.

\section*{Declarations}

\textbf{Availability of data and materials}
The source code for the Laplacian Graph Neural Process (Lap-GNP) models and all experiments presented in this study is publicly available on GitHub at \url{https://github.com/yzz980314/MetaMolGen}. The DrugBank dataset used for evaluation is a publicly accessible resource. All data preprocessing was performed using the RDKit library.

\textbf{Competing interests}
The authors declare that they have no competing interests.

\textbf{Funding}
This work was supported by the National Natural Science Foundation of China [61773020] and the Graduate Innovation Project of National University of Defense Technology [XJQY2024065].

\textbf{Authors' contributions}
Z.Y. conceived the methodology, developed the software, conducted the experiments, and wrote the original draft. Z.X. supervised the project, acquired funding, and reviewed \& edited the manuscript.

\textbf{Acknowledgements}
The authors would like to express their sincere gratitude to all the referees for their careful reading and insightful suggestions.

\bibliographystyle{IEEEtran} 
\bibliography{references} 

@inproceedings{gilmer2017neural,
  title={Neural message passing for quantum chemistry},
  author={Gilmer, Justin and Schoenholz, Samuel S and Riley, Patrick F and Vinyals, Oriol and Dahl, George E},
  booktitle={International conference on machine learning},
  pages={1263--1272},
  year={2017},
  organization={Pmlr}
}

@inproceedings{ying2018graph,
  title={Graph convolutional neural networks for web-scale recommender systems},
  author={Ying, Rex and He, Ruining and Chen, Kaifeng and Eksombatchai, Pong and Hamilton, William L and Leskovec, Jure},
  booktitle={Proceedings of the 24th ACM SIGKDD international conference on knowledge discovery \& data mining},
  pages={974--983},
  year={2018}
}

@article{kipf2017semi,
  title={Semi-supervised classification with graph convolutional networks},
  author={Kipf, TN},
  journal={arXiv preprint arXiv:1609.02907},
  year={2016}
}

@inproceedings{garnelo2018conditional,
  title={Conditional neural processes},
  author={Garnelo, Marta and Schwarz, Johannes and Rosenbaum, Dan and Viola, Fabio and Rezende, Danilo J and Eslami, S. M. Ali and Teh, Yee Whye},
  booktitle={International Conference on Machine Learning (ICML)},
  pages={1704--1713},
  year={2018},
  organization={PMLR}
}

@misc{yan2025multiscalegraphneuralprocess,
      title={A Multi-Scale Graph Neural Process with Cross-Drug Co-Attention for Drug-Drug Interactions Prediction}, 
      author={Zimo Yan and Jie Zhang and Zheng Xie and Yiping Song and Hao Li},
      year={2025},
      eprint={2509.15256},
      archivePrefix={arXiv},
      primaryClass={cs.LG},
      url={https://arxiv.org/abs/2509.15256}, 
}

@article{kim2019attentive,
  title={Attentive neural processes},
  author={Kim, Hyunjik and Mnih, Andriy and Schwarz, Jonathan and Garnelo, Marta and Eslami, Ali and Rosenbaum, Dan and Vinyals, Oriol and Teh, Yee Whye},
  journal={arXiv preprint arXiv:1901.05761},
  year={2019}
}

@inproceedings{xu2019how,
  title={How powerful are graph neural networks?},
  author={Xu, Keyulu and Hu, Weihua and Leskovec, Jure and Jegelka, Stefanie},
  booktitle={International Conference on Learning Representations (ICLR)},
  year={2019}
}

@inproceedings{morris2019weisfeiler,
  title={Weisfeiler and leman go neural: Higher-order graph neural networks},
  author={Morris, Christopher and Ritzert, Martin and Fey, Matthias and Hamilton, William L and Lenssen, Jan Eric and Rattan, Gaurav and Grohe, Martin},
  booktitle={Proceedings of the AAAI conference on artificial intelligence},
  volume={33},
  number={01},
  pages={4602--4609},
  year={2019}
}

@article{dwivedi2020benchmarking,
  title={Benchmarking graph neural networks},
  author={Dwivedi, Vijay Prakash and Joshi, Chaitanya K and Luu, Anh Tuan and Laurent, Thomas and Bengio, Yoshua and Bresson, Xavier},
  journal={Journal of Machine Learning Research},
  volume={24},
  number={43},
  pages={1--48},
  year={2023}
}

@inproceedings{kreuzer2021rethinking,
  title={Rethinking graph transformers with spectral attention},
  author={Kreuzer, Devin and Beaini, Dominique and Hamilton, Will and Liptchinsky, V and Pouliot, R},
  booktitle={Advances in Neural Information Processing Systems (NeurIPS)},
  volume={34},
  pages={22635--22647},
  year={2021}
}

@book{Chung1997Spectral,
  author    = {Fan R. K. Chung},
  title     = {Spectral Graph Theory},
  series    = {CBMS Regional Conference Series in Mathematics},
  volume    = {92},
  publisher = {American Mathematical Society},
  year      = {1997}
}

@book{LevinPeresWilmer2009MCMT,
  author    = {David A. Levin and Yuval Peres and Elizabeth L. Wilmer},
  title     = {Markov Chains and Mixing Times},
  publisher = {American Mathematical Society},
  year      = {2009},
  edition   = {1st}
}

@book{Norris1998Markov,
  author    = {J. R. Norris},
  title     = {Markov Chains},
  publisher = {Cambridge University Press},
  year      = {1998}
}

@book{LyonsPeres2016PTN,
  author    = {Russell Lyons and Yuval Peres},
  title     = {Probability on Trees and Networks},
  publisher = {Cambridge University Press},
  year      = {2016}
}

@inproceedings{hamilton2017inductive,
  title={Inductive Representation Learning on Large Graphs},
  author={Hamilton, William L. and Ying, Rex and Leskovec, Jure},
  booktitle={Advances in Neural Information Processing Systems (NeurIPS)},
  year={2017}
}

@inproceedings{velickovic2018graph,
  title={Graph Attention Networks},
  author={Veli{\v{c}}kovi{\'c}, Petar and Cucurull, Guillem and Casanova, Arantxa and Romero, Adriana and Li{\`o}, Pietro and Bengio, Yoshua},
  booktitle={International Conference on Learning Representations (ICLR)},
  year={2018}
}

@article{wu2020comprehensive,
  title={A Comprehensive Survey on Graph Neural Networks},
  author={Wu, Zonghan and Pan, Shirui and Chen, Fengwen and Long, Guodong and Zhang, Chengqi and Yu, Philip S.},
  journal={IEEE Transactions on Neural Networks and Learning Systems},
  volume={32},
  number={1},
  pages={4--24},
  year={2021},  
  publisher={IEEE}
}

@inproceedings{corso2020pna,
  title={Principal Neighbourhood Aggregation for Graph Nets},
  author={Corso, Gabriele and Cavalleri, Luca and Beaini, Dominique and Li{\`o}, Pietro and Veli{\v{c}}kovi{\'c}, Petar},
  booktitle={Advances in Neural Information Processing Systems (NeurIPS)},
  year={2020}
}

@inproceedings{li2018deeper,
  title={Deeper Insights into Graph Convolutional Networks for Semi-Supervised Learning},
  author={Li, Qimai and Han, Zhichao and Wu, Xiao-Ming},
  booktitle={AAAI Conference on Artificial Intelligence (AAAI)},
  year={2018}
}

@inproceedings{oono2020graph,
  title={Graph Neural Networks Exponentially Lose Expressive Power for Node Classification},
  author={Oono, Kenta and Suzuki, Taiji},
  booktitle={International Conference on Learning Representations (ICLR)},
  year={2020}
}

@inproceedings{alon2021bottleneck,
  title={On the Bottleneck of Graph Neural Networks and its Practical Implications},
  author={Alon, Uri and Yahav, Eran},
  booktitle={International Conference on Learning Representations (ICLR)},
  year={2021}
}

@inproceedings{topping2022understanding,
  title={Understanding Over-Squashing and Bottlenecks on Graphs},
  author={Topping, James and Di Giovanni, Francesco and Chamberlain, Benjamin P. and Dong, Xiaowen and Bronstein, Michael M.},
  booktitle={International Conference on Learning Representations (ICLR)},
  year={2022}
}

@inproceedings{bevilacqua2022equivariant,
  title={Equivariant Subgraph Aggregation Networks},
  author={Bevilacqua, Barbora and Zhou, Yifan and Li{\'o}, Pietro and Veli{\v{c}}kovi{\'c}, Petar and Barekatain, Mohammad},
  booktitle={International Conference on Learning Representations (ICLR)},
  year={2022}
}

@inproceedings{klicpera2019predict,
  title={Predict then Propagate: Graph Neural Networks meet Personalized PageRank},
  author={Klicpera, Johannes and Bojchevski, Aleksandar and G{\"u}nnemann, Stephan},
  booktitle={International Conference on Learning Representations (ICLR)},
  year={2019}
}

@inproceedings{klicpera2019diffusion,
  title={Diffusion Improves Graph Learning},
  author={Klicpera, Johannes and Wei{\ss}enberger, Stefan and G{\"u}nnemann, Stephan},
  booktitle={Advances in Neural Information Processing Systems (NeurIPS)},
  year={2019}
}

@inproceedings{ying2021graphormer,
  title={Do Transformers Really Perform Bad for Graph Representation?},
  author={Ying, Chengxuan and Cai, Tianle and Luo, Shengjie and Zheng, Shuxin and Ke, Guolin and He, Di and Shen, Tie-Yan and Liu, Yanming},
  booktitle={Advances in Neural Information Processing Systems (NeurIPS)},
  year={2021},
  note={Commonly referred to as Graphormer}
}

@inproceedings{rampasek2022gps,
  title={Recipe for a General, Powerful, Scalable Graph Transformer},
  author={Ramp{\'a}{\v{s}}ek, Ladislav and Galkin, Mikhail and Dwivedi, Vijay Prakash and Luu, Anh Tuan and Wolf, Guy and Beaini, Dominique},
  booktitle={Advances in Neural Information Processing Systems (NeurIPS)},
  year={2022}
}

@inproceedings{lim2022signnet,
  title={Sign and Basis Invariant Networks for Spectral Graph Representation},
  author={Lim, Derek and Rusch, T. Konstantin and Huang, Yu and Tang, Felipe},
  booktitle={Advances in Neural Information Processing Systems (NeurIPS)},
  year={2022},
  note={If you cite SignNet/BasisNet, align to the exact venue/version you use}
}

@article{giuliari2024positional,
  title={Positional diffusion: Graph-based diffusion models for set ordering},
  author={Giuliari, Francesco and Scarpellini, Gianluca and Fiorini, Stefano and James, Stuart and Morerio, Pietro and Wang, Yiming and Del Bue, Alessio},
  journal={Pattern Recognition Letters},
  volume={186},
  pages={272--278},
  year={2024},
  publisher={Elsevier}
}

@inproceedings{lee2020bootstrapping,
  title={Bootstrapping Neural Processes},
  author={Lee, Jaehoon and Teshima, Takeshi and Heo, Jayoung and Kim, Eunho and Kim, Sung Ju},
  booktitle={Advances in Neural Information Processing Systems (NeurIPS)},
  year={2020}
}

@inproceedings{kipf2016variational,
  title={Variational Graph Auto-Encoders},
  author={Kipf, Thomas N. and Welling, Max},
  booktitle={NeurIPS Workshop on Bayesian Deep Learning},
  year={2016}
}

@inproceedings{hasanzadeh2020bayesian,
  title={Bayesian graph neural networks with adaptive connection sampling},
  author={Hasanzadeh, Arman and Hajiramezanali, Ehsan and Boluki, Shahin and Zhou, Mingyuan and Duffield, Nick and Narayanan, Krishna and Qian, Xiaoning},
  booktitle={International conference on machine learning},
  pages={4094--4104},
  year={2020},
  organization={PMLR}
}

@inproceedings{percha2013profiling,
  title={Discovery and Explanation of Drug-Drug Interactions via Text Mining},
  author={Percha, Bethany and Altman, Russ B.},
  booktitle={Pacific Symposium on Biocomputing (PSB)},
  pages={410--421},
  year={2013}
}

@article{gottlieb2012ddi,
  title={INferring Drug Interactions from Multiscale Data},
  author={Gottlieb, Assaf and Stein, Gyuri Y. and Oron, Yitzhak and Ruppin, Eytan and Sharan, Roded},
  journal={Journal of the American Medical Informatics Association (JAMIA)},
  volume={19},
  number={5},
  pages={783--789},
  year={2012},
  publisher={Oxford University Press}
}

@article{ryu2018deepddi,
  title={Deep Learning Improves Prediction of Drug–Drug Interactions},
  author={Ryu, Jae Yong and Kim, Hyun Uk and Lee, Sang Yup},
  journal={Proceedings of the National Academy of Sciences (PNAS)},
  volume={115},
  number={18},
  pages={E4304--E4311},
  year={2018}
}

@article{zitnik2018decagon,
  title={Modeling Polypharmacy Side Effects with Graph Convolutional Networks},
  author={Zitnik, Marinka and Agrawal, Monica and Leskovec, Jure},
  journal={Bioinformatics},
  volume={34},
  number={13},
  pages={i457--i466},
  year={2018}
}

@article{law2014drugbank,
  title={DrugBank 4.0: Shedding New Light on Drug Metabolism},
  author={Law, Vivian and Knox, Craig and Djoumbou, Yifeng and Jewison, Tim and others},
  journal={Nucleic Acids Research},
  volume={42},
  number={D1},
  pages={D1091--D1097},
  year={2014}
}

@article{ma2023dualgnn,
  title={A dual graph neural network for drug--drug interactions prediction based on molecular structure and interactions},
  author={Ma, Mei and Lei, Xiujuan},
  journal={PLOS Computational Biology},
  volume={19},
  number={1},
  pages={e1010812},
  year={2023},
  publisher={Public Library of Science San Francisco, CA USA}
}

@article{liu2021comparative,
  title={A comparative study of graph neural networks for drug-drug interaction prediction},
  author={Liu, Z and Zhang, S},
  journal={IEEE Trans Biomed Eng},
  volume={38},
  number={5},
  year={2021}
}

@inproceedings{lin2020kgddi,
  title={KGNN: Knowledge graph neural network for drug-drug interaction prediction.},
  author={Lin, Xuan and Quan, Zhe and Wang, Zhi-Jie and Ma, Tengfei and Zeng, Xiangxiang},
  booktitle={IJCAI},
  volume={380},
  pages={2739--2745},
  year={2020}
}

@article{li2024multimodal,
  title={Deep learning for drug-drug interaction prediction: A comprehensive review},
  author={Li, Xinyue and Xiong, Zhankun and Zhang, Wen and Liu, Shichao},
  journal={Quantitative Biology},
  volume={12},
  number={1},
  pages={30--52},
  year={2024},
  publisher={Wiley Online Library}
}

@article{Nyamabo2021ssidddi,
  title={SSI--DDI: substructure--substructure interactions for drug--drug interaction prediction},
  author={Nyamabo, Arnold K and Yu, Hui and Shi, Jian-Yu},
  journal={Briefings in Bioinformatics},
  volume={22},
  number={6},
  pages={bbab133},
  year={2021},
  publisher={Oxford University Press}
}

@article{Shen2025baseline,
  title={Benchmarking drug-drug interaction prediction methods: a perspective of distribution changes},
  author={Shen, Zhenqian and Zhou, Mingyang and Zhang, Yongqi and Yao, Quanming},
  journal={Bioinformatics},
  pages={btaf569},
  year={2025},
  publisher={Oxford University Press}
}

@misc{yan2025metamolgenneuralgraphmotif,
      title={MetaMolGen: A Neural Graph Motif Generation Model for De Novo Molecular Design}, 
      author={Zimo Yan and Jie Zhang and Zheng Xie and Chang Liu and Yizhen Liu and Yiping Song},
      year={2025},
      eprint={2504.15587},
      archivePrefix={arXiv},
      primaryClass={cs.LG},
      url={https://arxiv.org/abs/2504.15587}, 
}

@inproceedings{paszke2019pytorch,
  title={PyTorch: An Imperative Style, High-Performance Deep Learning Library},
  author={Paszke, Adam and Gross, Sam and Massa, Francisco and Lerer, Adam and Bradbury, James and Chanan, Gregory and Killeen, Trevor and Lin, Zeming and Gimelshein, Natalia and Antiga, Luca and others},
  booktitle={Advances in Neural Information Processing Systems},
  pages={8026--8037},
  year={2019}
}

@inproceedings{fey2019fast,
  title={Fast Graph Representation Learning with PyTorch Geometric},
  author={Fey, Matthias and Lenssen, Jan Eric},
  booktitle={ICLR Workshop on Representation Learning on Graphs and Manifolds},
  year={2019}
}

@article{li2020distance,
  title={Distance encoding: Design provably more powerful neural networks for graph representation learning},
  author={Li, Pan and Wang, Yanbang and Wang, Hongwei and Leskovec, Jure},
  journal={Advances in Neural Information Processing Systems},
  volume={33},
  pages={4465--4478},
  year={2020}
}

\clearpage

\tableofcontents
\clearpage

\end{document}